\documentclass[accepted]{uai2025} % for initial submission
\usepackage[american]{babel}
\usepackage{natbib} % has a nice set of citation styles and commands
\usepackage{mathtools} % amsmath with fixes and additions
\usepackage{booktabs} % commands to create good-looking tables
\usepackage{tikz} % nice language for creating drawings and diagrams

\usepackage{amsthm}
\newtheorem{theorem}{Theorem}
\usepackage{algorithm}
\usepackage{algpseudocode}
\usepackage{amsthm}
\usepackage{amsfonts}
\usepackage{multirow}

\newcommand{\Var}{\text{Var}}

\newcommand{\cN}{\mathcal{N}}

\newcommand{\E}{\mathbb{E}}
\newcommand{\Prob}{\mathbb{P}}
\newcommand{\barsig}{\bar{\sigma}_{1j}}
\newcommand{\barmu}{\bar{\mu}_{1j}}
\newcommand{\bareta}{\bar{\eta}_{1j}}
\DeclareMathOperator*{\argmin}{arg\,min}
\DeclareMathOperator*{\argmax}{arg\,max}

\title{Statistical Significance of Feature Importance Rankings}

\author[1]{\href{mailto:<jeremy\_goldwassser@berkeley.edu>?Subject=Your UAI 2025 paper}{Jeremy Goldwasser}{}}
\author[2]{Giles Hooker}

\affil[1]{Department of Statistics\\ 
University of California, Berkeley}
\affil[2]{Department of Statistics and Data Science\\
University of Pennsylvania}

\begin{document}
\maketitle

\begin{abstract}
  Feature importance scores are ubiquitous tools for understanding the predictions of machine learning models. 
  However, many popular attribution methods suffer from high instability due to Monte Carlo sampling.
  Leveraging novel ideas from hypothesis testing, we devise techniques that ensure the most important features are correct with high-probability guarantees.
  These are capable of assessing both the set of $K$ top-ranked features as well as the order of its elements.
  Given local or global importance scores, we demonstrate how to retrospectively verify the stability of the highest ranks.
  We then introduce two efficient sampling algorithms that identify the $K$ most important features, perhaps in order, with bounded error rate.
  The theoretical justification for these procedures is validated empirically on SHAP and LIME.
\end{abstract}

\section{Introduction}\label{sec:intro}

Many machine learning (ML) algorithms have impressive predictive power but poor interpretability relative to simpler alternatives like decision trees and linear models. This tradeoff has motivated a wide body of work seeking to explain how black-box models make predictions \citep{XAI}. Such work is essential for building trust in ML systems in areas like finance, healthcare, and criminal justice, in which the consequences of model misbehavior may be severe \citep{ML_finance, ML_healthcare, ML_crime, ML_crime2}. Interpretable ML methods can also help develop understanding of complex processes, augmenting domain knowledge with new hypotheses. 

To that end, feature importance scores quantify how much the features contribute to the model's predictions. 
These may explain model behavior at the resolution of an individual input (\textit{local}) or in aggregate (\textit{global}).
A wide range of methods have been proposed, detailed in Section \ref{sec:background}. 

% In this paper, we argue that the 
The particular value of a feature's importance score may be of less practical interest than its ranking: which features are the most important, and the order of these highlighted features. %relative
A scientist using a machine learning model to predict disease risk from a patient’s genetic profile will focus on the genes with the highest importance scores for further study, prioritizing the ranking of the metric over its specific values.
Similarly, explanations often report only a small number of features in order not to overwhelm the user. %of individual predictions
LIME \citep{LIME}, for example, explicitly regularizes to report a fixed number of features. 
%, with subtle definitions and differences between them

\begin{figure*}
    \centering
    \includegraphics[width=0.7\textwidth]{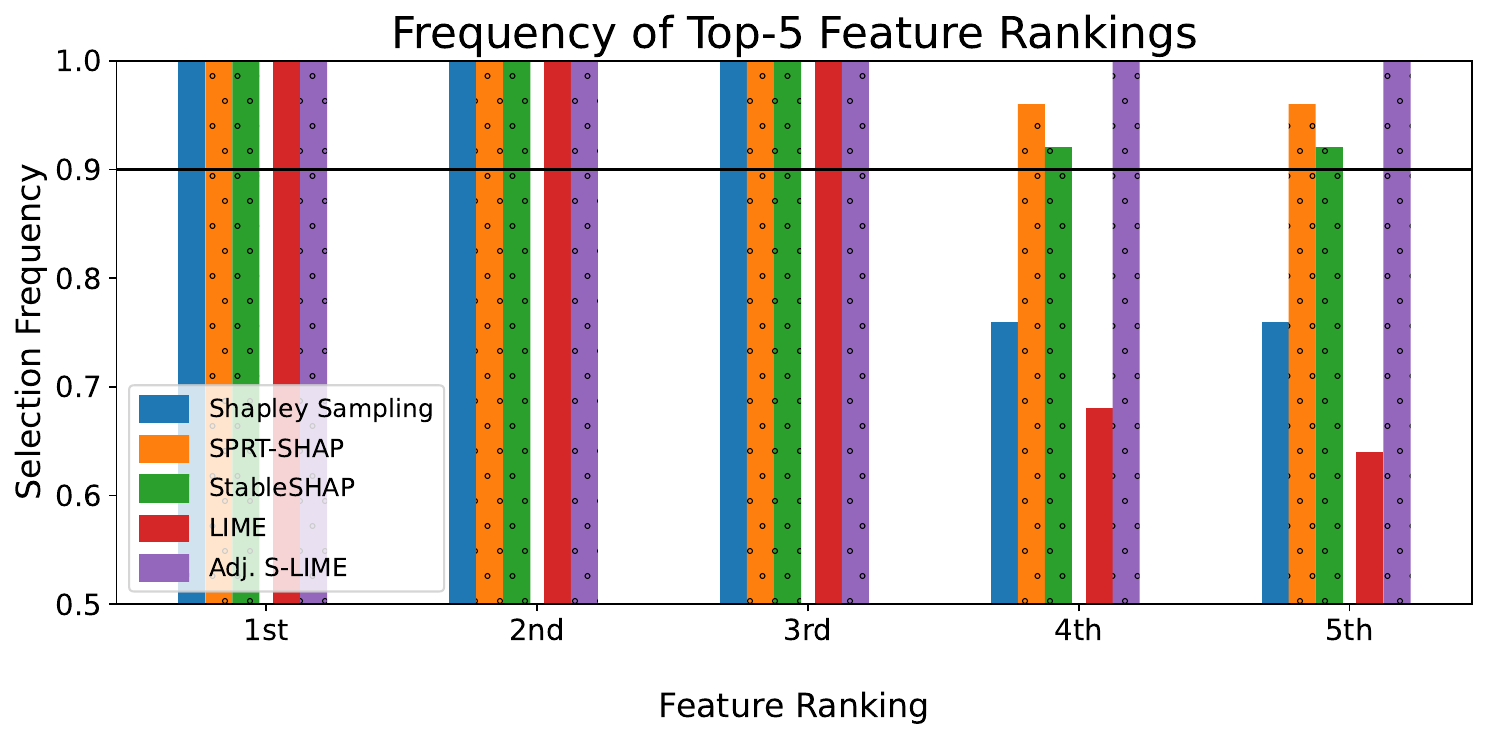}
    \caption{Instability of top-5 feature rankings, Adult Census Income dataset. StableSHAP, SPRT-SHAP, and Adj. S-LIME are our contributions, run at $\alpha=0.1$; meaning 90\% of replicates should return the same feature set. The computational budget of StableSHAP and Shapley Sampling are the same.}
    \label{fig:instability}
\end{figure*}

Unfortunately, many standard attribution methods suffer from instability induced by random sampling. 
Rerunning the same procedure could yield different explanations for which features are the most important. 
% For local methods such as SHAP and LIME, obtaining scores explicitly requires Monte Carlo sampling. 
Local methods such as SHAP and LIME require Monte Carlo sampling to calculate scores \citep{SHAP}. 
Global scores like LOCO are calculated on some finite set of input data, yet are used to infer across the entire population of data \citep{loco}.
% A straightforward approach for global attributions is to average local attributions across some number of inputs, e.g. LOCO \citep{loco}, permutation importance \citep{RFs, rudin}, and global aggregations of local measures \cite{SAGE, treeSHAP, global_aggregations}.
This inherent lack of reproducibility seriously undermines the credibility of these analyses \citep{PCS}.
% A number of adjustments have been shown to reduce the variability of local attribution values (e.g. KernelSHAP \citep{SHAP}). 
% % however, they do not necessarily stabilize the more salient property of the features' rankings. 
% However, substantial instability remains, including with respect to the features' rankings.

To address this, we cast a large class  of local and global attributions into a unifying framework, based on properties of unbiasedness and asymptotic normality. 
For any method in this framework, we propose techniques to verify that the observed importance rankings are correct with high-probability guarantees.
Amongst the highest-ranking features, these techniques assess the stability of both the top-$K$ set as well as their ordering relative to one another.

We provide retrospective tools that analyze the rankings of given feature attributions (Section~\ref{sec:retro}).
Then, we propose two sampling methods which ensure the $K$ highest-ranked features are correct with probability exceeding $1-\alpha$ (Section~\ref{sec:algorithms}). 
The first of these methods leverages the Sequential Probability Ratio Test (SPRT) from \citet{Wald} to sample continually. 
The second approach, more efficient for certain scores, iterates sample size calculations on ambiguously-ranked pairs. 
Applied to Shapley values, we refer to these algorithms as SPRT-SHAP and StableSHAP.
Figure \ref{fig:instability} highlights the improvement in stability of our sampling algorithms relative to baselines on SHAP and LIME.
% INCLUDE LINK.
\footnote{Our Python code and experimental results are at \url{https://github.com/jeremy-goldwasser/feature-rankings}.}

\section{Feature Importance Scores}\label{sec:background}

% This section surveys popular local and global attribution methods. All discussed scores are either sample averages or linear regression coefficients.

\subsection{Local Feature Importance}

A broad range of research has addressed the question of how best to attribute variable importance for individual predictions. 
% Many methods are specific to the particular choice of model  \citep{RFs, FIR, integrated_gradients}, whereas others are model-agnostic \citep{loco, rudin}.
SHAP \citep{SHAP} and LIME \citep{LIME} are amongst the most popular local attribution methods. 
Both are model-agnostic and entail random Monte Carlo computation.
We provide a brief review of these two techniques, then introduce global feature importance.

\subsubsection{SHAP}
SHAP (SHapley Additive exPlanations) is a special case of the Shapley value from game theory \citep{shapley_original}. 
Shapley values, and SHAP by extension, uniquely satisfy several reasonable desiderata for credit allocation. %to ``players'' in a cooperative game. 
% In the context of ML feature importance, players are features, and the game is prediction. 
%``players'' (i.e. features)
% Let $d$ be the number of variables to score, and let 
% They average the marginal contribution of each variable relative to all subsets of the other features.
For any subset $S$ of the $d$ variables, define the value function \mbox{$v:S\rightarrow\mathbb{R}$} and weighting kernel $w_S~=~\binom{d-1}{d-\vert S \vert -1}^{-1}$. The Shapley value for variable $j$ is
% Let the value function $v:S\rightarrow\mathbb{R}$ evaluates some subset $S$ of the $d$ variables. 
% The Shapley value for variable $j$ averages its marginal contribution to the value function relative to all subsets of the other features.

% \begin{align}\label{eq:Shapley Value}
%     % \phi_j(v) &:= \frac{1}{d!} \sum_{\pi \in \Pi(d)}  v(S_j^\pi \cup \{j\}) - v(S_j^\pi) \\
%      \phi_j(v) &= \frac{1}{d}\sum_{S \subseteq [d]\backslash\{j\}} w_S (v(S \cup \{j\}) - v(S)).%\nonumber
% \end{align}
\begin{equation}\label{eq:Shapley Value}
     \phi_j(v) = \frac{1}{d}\sum_{S \subseteq [d]\backslash\{j\}} w_S (v(S \cup \{j\}) - v(S)).
\end{equation}
SHAP popularized the use of Shapley values for local feature attribution. In this context, $v(S)$ is the prediction the model $\hat{f}$ would have made on input $x$ if it only had access to the features $x_S$ with indices in $S \subseteq [d]$.
In theory, this prediction may be obtained by refitting the model on those features \citep{strumbelj_kononenko09, lipotevsky}.
For computational convenience, however, it is common to sample the features in its complement $S^c$, concatenate them onto $x_S$, and take the prediction $\hat{f}(x_S, x_{S^c})$. 
The unknown features are usually sampled from their marginal distributions, as in SHAP \citep{SHAP, strumbelj_kononenko14}. Other works propose using conditional distributions $X_{S^c} | X_S$  instead \citep{aas, frye}.
% The statistical guarantees of our methods hold under any choice of $v(S)$, i.e. for any Shapley value.

Computing the exact Shapley value \eqref{eq:Shapley Value} requires evaluating $v(S)$ for $O(2^d)$ terms. 
When $d$ is large, this is computationally prohibitive, so approximation algorithms must be used instead.
The basic approach is Shapley Sampling (Algorithm \ref{alg:SS}) \citep{strumbelj_kononenko14}.
Rather than using all subsets of $[d]\backslash\{j\}$, this algorithm samples $n$ subsets at random. 
\begin{equation}\label{eq:Shapley Sampling}
    \hat{\phi}_j(v) = \frac{1}{n} \sum_{i=1}^n v(S_j^i \cup \{j\}) - v(S_j^i).
\end{equation}
This sample average is unbiased for the true Shapley value, meaning $\E[\hat\phi_j]=\phi_j$ \citep{strumbelj_kononenko09}. 
\citet{SHAP} later proposed KernelSHAP, a more efficient estimator than Shapley Sampling. 
KernelSHAP obtains all $d$ Shapley values at once in a linear regression framework.
% \citet{CovertLee} proved its bias is negligible. 

\subsubsection{LIME}\label{sec:LIME}

Local Interpretable Model-agnostic Explanations, or LIME, explains an individual prediction via a local surrogate \citep{LIME}. Around the input in question, LIME first samples a large number of random data points. 
These are passed through the model, producing a collection of labeled pseudodata. 
The pseudodata is then used to fit an inherently interpretable model, which summarizes the salient factors driving the original model's prediction. 

The explanation model is typically a linear model, possibly regularized for smoothness or sparsity. For example, the default choice in the \texttt{lime} Python package is a sparse linear model with $K$ nonzero coefficients. 
% In this case of a linear model, the regression coefficients provide natural feature importance scores. 
Because linear models associate each feature with a unique regression coefficient, they provide a natural framework for feature attributions. 
% In general, linear models 

In this setting, LIME importance rankings may be obtained in a number of ways. One could score each feature as the product of its regression coefficient and observed value, with the $K$ highest scores corresponding to the most important features. 
% To rank the $K$ most important features, one could choose the $K$ highest products of OLS coefficient and original feature value.
When the explanation model is sparse, another option is to take the first $K$ features that enter the Lasso path with decreasing $\lambda$ \citep{LASSO, Tibshirani_2011}.
Finally, the authors propose $K$-Lasso, ranking the top $K$ features with the order they enter the Least Angle Regression path \citep{LARS}. 
% In this context, the ranking could be the order the features are selected, rather than their coefficient values.

\subsection{Global Feature Importance}

A number of methods summarize a model's general behavior with global feature importance scores. 
The global scores $\hat\phi_j\;\forall j$ are calculated from a finite number of inputs $x_i$. 
They may be used as a proxy for $\phi_j$, the population score describing the model across the entire sample space $\mathcal{X}$.
% Their purpose may be to describe the model across entire population of input samples $\mathcal{X}$. 

Most global attribution methods are a sample average of scores on individual data points. 
For local score $s_j^i$ on feature $j$ of input $i$, these attributions are 
% $\hat\phi_j = \frac{1}{n}\sum_{i=1}^n s_j^i$.
% When the inputs used to compute $\hat\phi_j$ are selected at random from $\mathcal{X}$, $\hat\phi_j$ is unbiased for the population score $\phi_j$. 
\begin{equation}\label{eq:global_avg}
    \hat\phi_j = \frac{1}{n}\sum_{i=1}^n s_j^i.
\end{equation}
One approach merely takes the average of local feature attribution scores. 
For example, SAGE is the mean LossSHAP score; this is a Shapley value based on the per-sample loss, rather than the prediction. 
A global Shapley method that does not require labels is the average absolute SHAP value \citep{SHAP, treeSHAP}.
Similarly, \citep{LIME} computes the mean absolute LIME score to present global explanations. 
Analogous strategies have been suggested for counterfactual explanations \citep{ley2022globalcounterfactuals}.
More broadly, \citet{global_aggregations} proposes global importance scores that average the absolute value of any local metric.

Another subset of global scores taking the form \eqref{eq:global_avg} compares model performance before and after controlling for feature $j$. 
The performance is evaluated with some loss function $\mathcal{L}$ on labeled data points $\{(x_i, y_i)\}_{i=1}^n$. 
These methods modify either the data or model to produce some altered prediction $\tilde{y}_{ij}$.
Their local scores take the form
$$s_j^i = \mathcal{L}(y_i, \hat{y}_i)-\mathcal{L}(y_i, \tilde{y}_{ij}).$$
Here, we survey choices of $\tilde{y}_{ij}$ for a number of popular loss-based frameworks.
\begin{itemize}
    \item \textbf{Permutation Importance} \citep{RFs, rudin}. Averaging across unseen data points, this approach takes predictions after permuting the values of feature $j$. Its original formulation, Mean Decrease in Accuracy (MDA), used out-of-bag training inputs in random forests.
    \item \textbf{Conditional Variable Importance} \citep{Strobl2008Conditional}. Permutation Importance samples features from their marginal distribution, producing unlikely data with potentially inaccurate predictions. Conditional Variable Importance instead samples each feature from its distribution conditional on the other features.
    \item \textbf{Leave-One-Covariate-Out} \citep{loco, verdinelli2023featureimportancecloserlook}. LOCO retrains the model without each feature $j$, then predicts $\tilde{y}_{ij}$ using all other features. 
    \item \textbf{Permute-and-relearn}. This technique retrains after imputing random values of a given feature from its marginal \citep{hooker2021unrestricted} or conditional \citep{Mentch2016Quantifying} distribution. 
\end{itemize}
% A large category of such methods are loss-based

A separate strategy trains an interpretable surrogate model on the predictions of the black-box model \citep{molnar2022, distillation}. When a linear model is used, its regression coefficients may be taken as importance scores in the same fashion as LIME. 
% Assuming normal errors, Ordinary Least Squares coefficients are unbiased, and their variance can be estimated with canonical formulas.

% \section{Statistical Tools}
% In this section we review statistical techniques that will prove useful to stabilize feature importance scores. 

\section{Rank Verification Review}

\subsection{Stable Importance Rankings}\label{sec:background-stability}

A small body of work seeks to identify the most important features from estimated feature importances. 
\citet{benjamini} established population rankings in retrospect via simultaneous confidence intervals, adjusted for multiple testing with procedures such as Holm's method \citep{holm1979simple}. 
While valid at level $\alpha$, the use of such multiple testing corrections drastically reduces the power of this procedure. 
The p-values are inflated by a factor of $O(d^2)$, 
the total number of pairwise comparisons.
%the number of pairwise comparisons necessary to verify the rankings.

% \citet{TopSHAP}, \citet{sharp}, 
% \citet{top_k_shapley_graph}, \citet{topKbandit}, and \citet{shap@k} propose sampling algorithms designed to stabilize the set of $K$ highest Shapley values. 
% The latter two 
A number of works propose sampling algorithms designed to stabilize the set of $K$ highest Shapley values \citep{top_k_shapley_graph, sharp, TopSHAP, topKbandit, shap@k}.
The latter two present probabilistic guarantees for their top-$K$ bandit algorithms, albeit in terms of unknown parameters based on the gaps between Shapley values.
\citet{shap@k} only guarantees $\epsilon$-approximate solutions --- allowing incorrect top-$K$ features to be included, so long as they are within $\epsilon$ of the true top $K$. 
The algorithms are further hindered by the use of loose bounds like the Bonferroni correction.

%a naive and bandit
% \citet{topKbandit} analyzes conditions under which two algorithms stabilize the top-$K$ set with high probability. Both require the sample budget to exceed a certain unknown quantity, based on the gap between the $K$ and $K+1$\textsuperscript{th} true Shapley value.
% \citet{topKbandit}  
% analyzes conditions under which two algorithms stabilize the top-$K$ set with high probability. Both require the sample budget to exceed a certain unknown quantity, based on the gap between the $K$ and $K+1$\textsuperscript{th} true Shapley value.

% Similarly, \citet{shap@k} ensures its bandit algorithms yield the correct top-$K$ set with high probability assuming these unknown gaps are not too small. 
% Like \citet{benjamini}, these algorithms use the Bonferroni correction to account for multiple testing, hindering the runtime of their algorithms.
% % these algorithms are also hindered by multiple testing corrections.
% They also target the weaker condition of $\epsilon$-approximate solutions --- allowing incorrect top-$K$ features to be included, so long as they are within $\epsilon$ of the true top $K$. 

% demonstrates that two algorithms stabilize the top-$K$ set with high probability if the sample budget exceeds a certain quantity; 

\subsection{Gaussian Methodology}\label{sec:goldwasser}

Given a set of random variables, it is often of interest to verify whether the highest observed value - or values - indeed matches the population ranking with high probability. 
The methods discussed in \ref{sec:background-stability} do so coarsely, relying on multiple testing adjustments, loose bounds, and unknown parameters. 
In contrast, more powerful verification methods have been studied under various probability distributions via selective inference \citep{fithian_hung, PNAS}.

% Recently, \citet{goldwasser2025gaussianrankverification} provided the first selective-inference based rank verification for Gaussian data with unequal variances.
\citet{goldwasser2025gaussianrankverification} introduced the first selective inference-based methods to verify the ranks of Gaussian data with unequal variances.
For $j \in [d]=\{1,\ldots, d\}$, consider independent random variables $X_j \sim \cN(\mu_j, \sigma_j^2)$, where $\sigma_j$ is known and $\mu_j$ is unknown. Let $x_j$ denote the realized values of these random variables $X_j$. 
For notational convenience, sort the variables according to their order statistics, such that $X_1$ has the highest observed value $x_1$, $X_2$ is second-largest, etc. 

The primary objective is to verify the ``winner'' $X_1$ as the ``best,'' meaning $\mu_1 > \mu_j$ for all $j > 1$. 
Its rank is verified upon rejecting the null hypothesis $H_{01}$ that $\mu_1$ is \textit{not} the highest, conditioned on the selection event that $X_1$ wins.%was higher than all other observed $X_k$.
\begin{equation}\label{eq:nullH01}
    H_{01}: \bigcup_{j>1}\underbrace{ \bigl\{\mu_1 \leq  \mu_j \ \vert \ X_1 > \max_{k>1} X_k\bigr\}}_{H_{01j}}.
\end{equation}
To test this null, define \mbox{$\barmu = \frac{\sigma_j^2 x_1 + \sigma_1^2 x_j}{\sigma_1^2 + \sigma_j^2}$}, \mbox{$\barsig^2 = \frac{\sigma_1^4}{\sigma_1^2+\sigma_j^2}$}, and \mbox{$\bareta=\max(\barmu, \max_{k\neq 1,j} x_k)$}. 
This parameterizes a normal distribution with mean $\barmu$ and variance $\barsig^2$, whose left tail has been truncated at $\bareta$. 
For standard normal CDF $\Phi(\cdot)$, further define $p_{1j}$: 
% For standard normal CDF $\Phi(\cdot)$, further define $p_{1j}$, the right tail of a normal distribution with left truncation at $\bareta$:
\begin{equation}\label{eq:pval}
    p_{1j} = \frac{1-\Phi(\frac{x_1 - \barmu}{\barsig})}{1-\Phi(\frac{\bareta - \barmu}{\barsig})}.
\end{equation}
% \noindent where $\Phi(\cdot)$ is the CDF of the standard normal distribution. 

$p_{1j}$ is the tail mass of this truncated normal distribution above $x_1$.
% Theorem 1 from 
\citet{goldwasser2025gaussianrankverification} shows it is a valid p-value for $H_{01j}$ \eqref{eq:nullH01}. Furthermore, $H_{01}$ may be tested at level $\alpha$ with
\begin{equation}\label{eq:test}
    p_1^* = \max_{j>1} p_{1j}.
\end{equation}
The winner is verified when $p_1^*\leq\alpha$, rejecting $H_{01}$. This is equivalent to rejecting $H_{01j}$ with significant p-values $p_{1j}\leq\alpha$ for all $j$.

This result has useful extensions for top-$K$ rank verification, still without explicit multiple testing adjustments.
% Both extensions entail testing nulls $H_{0i} = \bigcup_{j>i} H_{0ij}$ in the same form as $H_{01}$, verifying $\mu_i$ for lower-ranked $i$. 
These verify all ranks $i\leq K$ by rejecting nulls of the form
\begin{equation}\label{eq:nullH0i}
    H_{0i}: \bigcup_{j} \underbrace{\bigl\{\mu_i \leq  \mu_j \ \vert \  X_i > \max_{k>i} X_k\bigr\}}_{H_{0ij}}.
\end{equation}
\textbf{Procedure 1 (Top-$K$ Ranks).} 
This outputs an integer $K\geq 0$ such that the probability that the top-$K$ rankings are correct is at least $1-\alpha$.
It iterates the test in Equation \eqref{eq:test} on successive ranks until a failure to reject (Alg. \ref{alg:procedure1}). 

% First, test $H_{01}$ to verify $\mu_1$ as the best with $p_1^* = \max_{j>1} p_{1j}$, i.e. testing $H_{01j}$ for all $j>1$.
% If this rejects at level $\alpha$, then test $H_{02}$ to verify $\mu_2$ as the second best with $p_{2j}\;\forall j>2$, again with Equation~\eqref{eq:pval} on $H_{02j}$.
First, test $H_{01}$ with $p_1^* = \max_{j>1} p_{1j}$.
If $p_1^*\leq \alpha$, then test $H_{02}=\bigcup_{j>2} H_{02j}$ with $p_2^* = \max_{j>2} p_{2j}$.
% to verify $\mu_2$ as the second best with $p_{2j}\;\forall j>2$, again with Equation~\eqref{eq:pval} on $H_{02j}$.
Continuing for $K\geq 0$ rejections, 
% After $K\geq 0$ rejections have occurred, 
stop at the first failure to reject,  wherein \mbox{$p_{(K+1)j} > \alpha$} for some  $j>K+1$.

 \textbf{Procedure 2 (Top-$K$ Set)}. 
 This test evaluates whether the set of $K$ largest observed elements are guaranteed to have the highest means with probability exceeding $1-\alpha$ (Alg. \ref{alg:procedure2}). 
 Unlike in Procedure 1, here $K$ is fixed a priori, and the orderining within the top $K$ does not matter.

 % To do so, it tests whether p-values $p_{ij}$ \eqref{eq:pval} for $H_{0ij}$ are significant at level $\alpha$ for all $i\leq K$ and $j>K$ (Alg. \ref{alg:procedure2}). 
  % This verifies whether the elements in the top $K$ indeed have larger means than those in the bottom $d-K$.
 % Each null $\widetilde{H}_{0i}$ that $\mu_i$ is \textit{not} in the top $K$ is tested with
 % $$\tilde{p}_i^* = \max_{j>K} p_{ij}.$$
Procedure 2 tests $\widetilde{H}_{0i} = \bigcup_{j>K} H_{0ij}$, i.e. that $\mu_i$ is not actually in the top $K$ set, with \mbox{$\tilde{p}_i^* = \max_{j>K} p_{ij}$}. 
When all $K$ nulls $\widetilde{H}_{0i}$ reject, the top-$K$ set is verified.
This is equivalent to having $p_{ij}\leq\alpha$ for all  $i\leq K$ and $j>K$.

% We refer the reader to \citet{goldwasser2025gaussianrankverification} for the proofs.
Corollaries 1 and 2 in \citet{goldwasser2025gaussianrankverification} establish the validity of Procedures 1 and 2. We refer the reader to the original manuscript for the proofs.

\section{Retrospective Score Verification}\label{sec:retro}

Under mild assumptions, most of the scores discussed in Section~\ref{sec:background} are normally distributed and unbiased. Their variance can be well-approximated with simple techniques.%, computationally cheap techniques. 

To see this, consider first attributions of the form \mbox{$\hat\phi_j = \frac{1}{n}\sum_{i=1}^n s_j^i$}, such as Shapley Sampling and LOCO.
When the samples used to compute $\hat\phi_j$ are selected at random, $\hat\phi_j$ is unbiased for the population score $\phi_j$. 
Moreover, with sufficiently large $n$, the distribution of $\hat\phi_j$ converges to a normal distribution centered around $\phi_j$ by the central limit theorem. 
The sample variance $\hat\sigma_j^2 = \frac{1}{n-1}\sum_{i=1}^n (s_j^i-\bar{s}_j)^2$ is unbiased for $\sigma_j^2$, converging at an $O(n^{-1/2})$ rate.

In addition, the scores of local (LIME) and global surrogate methods are linear regression coefficients. 
Assuming normal errors, Ordinary Least Squares coefficients are unbiased, and their variance can be estimated with standard formulas.
Similarly, KernelSHAP expresses Shapley values as linear regression coefficients, solving with a weighted least squares. 
\citet{CovertLee} proved it is asymptotically normal with negligible bias. 
They also introduced a variance estimator, studied and improved upon by \citet{controlSHAP}.
Finally, LIME with $K$-Lasso selects features with scaled correlations that are asymptotically normal \citep{S-LIME}.

Further assume the importance scores $\hat\phi_j$ are independent. 
This certainly holds in some cases. 
For example, Shapley Sampling \eqref{eq:Shapley Sampling} uses different subsets $S_j^i$ for each feature, so the resulting attributions are independent. 
However, mild correlation may exist in methods like KernelSHAP, where the same data is used to estimate multiple scores. 
Nevertheless, our empirical results indicate the ensuing procedures are always valid, and in fact somewhat conservative.
Appendix \ref{apx:correlation} evaluates the merits of alternative approaches which use correlated testing. 

\textbf{Main result.} Let $\hat{\phi}_1,\ldots,\hat{\phi}_d$ be a set of such feature importance scores. %, or perhaps their absolute values. 
If the user desires, these may be absolute values. 
Assume they are independent, normal, unbiased, and with known oracle variance ($\hat\sigma_j^2 = \sigma_j^2$).
Then the procedures from \citet{goldwasser2025gaussianrankverification} may be applied to verify the observed importance rankings.

% Letting $(\hat{k})$ denote the $k$\textsuperscript{th} largest value,
Let $(\hat{k})$ denote the $k$\textsuperscript{th} largest score.%, perhaps sorting $\hat\phi$ by absolute value.
Procedure 1 (Alg. \ref{alg:procedure1}) yields some non-negative integer $K$ such that with probability at least $1-\alpha$,
$$\phi_{(\hat{1})} > \ldots > \phi_{(\hat{K})} > \max_{\ell>K} \phi_{(\hat{\ell})}.$$
% We would like to identify, at some confidence level, how many top-ranked features are indeed in the correct order. 
Procedure 2 (Alg. \ref{alg:procedure2}) tests the stability of the top-$K$ set, for any user-defined $K\in \{1,\ldots, d-1\}$. When its test rejects at level $\alpha$, the set of $K$ most important features is correct, again with the same high-probability guarantee.

The test \eqref{eq:test} may also be applied simultaneously, indicating whether the $k$\textsuperscript{th}-ranked feature is indeed higher than all lower ranks for all $k$ of interest. 
Doing so is akin to the $\texttt{lm}$ function in \texttt{R}, which tests the statistical significance of all coefficients in a linear model.
Not all significant results necessarily hold at level $\alpha$ due to multiple testing; 
nevertheless, this provides a concise summary of which ranks are likely stable.

% Without loss of generality,  one may choose to rank attributions by their absolute values. 
% We present this as an option in our code.

\section{Stabilized Top-$K$ Algorithms}\label{sec:algorithms}

This section introduces algorithms that guarantee the $K$ highest-ranking features, and perhaps their relative ordering, are correct with probability at least $1-\alpha$, where $K$ and $\alpha$ are predetermined by the user. 
% This section introduces novel sampling procedures that efficiently guarantee the top $K$ ranks are stable with probability exceeding $1-\alpha$, where $K$ and $\alpha$ are predetermined by the user. 
This is motivated by numerous applications in which the $K$ most important features are analyzed, e.g. \citet{top_k_shapley_graph,top_k_shapley_management,top_k_shapley_neuron}.
Again, features may be sorted via their absolute values.

The algorithms sample until Procedure 1 or 2 verifies the top-$K$ features. 
To guarantee validity, Section~\ref{sec:SPRT-SHAP} uses modified hypothesis tests based on the SPRT \citep{Wald}, whereas Section~\ref{sec:StableSHAP} performs sample size calculations to efficiently obtain each attribution.

\subsection{SPRT Approach}\label{sec:SPRT-SHAP}

A naive top-$K$ ranking strategy would draw samples until Procedure 1 from Section~\ref{sec:goldwasser} rejects for the $K$ highest feature importances. 
An analogous approach for the top-$K$ set samples until Procedure 2 rejects.
% As discussed in Section~\ref{sec:SPRT}, however, doing so would not necessarily control the error rate at level $\alpha$. 
However, doing so would not necessarily control the error rate at level $\alpha$.
This is because standard hypothesis tests like the p-values $p_{ij}$ from \citet{goldwasser2025gaussianrankverification} are not valid under \textit{optional stopping}, when data is accumulated until the moment it indicates a significant result. 
This process inflates the Type I error rate because it permits the data to be tested multiple times without adjustment.

To address this, we modify Procedures 1 and 2 so the tests they conduct are valid under optional stopping.
The canonical choice in this setting is the Sequential Probability Ratio Test \citep{Wald}. 
% The Sequential Probability Ratio Test is a valid test designed for the optional stopping setting \citep{Wald}. 
After any number of samples have been drawn, SPRT computes the likelihood ratio 
\begin{equation}\label{eq:SPRT}
    T = \frac{\max_{\theta \in H_1} \Prob(X\ \vert \ \theta)}{\max_{\theta \in H_0} \Prob(X\ \vert \ \theta)}.
\end{equation}
Set Type I and II error rates $\alpha$ and $\beta$. 
The test accepts $H_1$ when $T \geq \frac{1-\beta}{\alpha}$, accepts $H_0$ when $T \leq \frac{\beta}{1-\alpha}$, and continues sampling otherwise. 

In the context of feature importance rankings, we construct the SPRT likelihood ratio $T$ with the following theorem.
The proof, in Appendix \ref{apx:SPRT-proof}, involves maximum likelihood estimation in a selective inference framework. 

\begin{theorem}\label{thm:LR}
Assume $X_j \sim \cN(\mu_j, \sigma_j^2)$ independently, where $\sigma_j$ is known.
Let $\phi$ and $\Phi$ be the standard normal PDF and CDF, and recall the definitions of $\barmu$, $\barsig$, and $\bareta$ from Section~\ref{sec:goldwasser}.
For any Type I and II error thresholds $\alpha$ and $\beta$, a valid SPRT test statistic for \mbox{$H_{01j}:\mu_1 \leq \mu_j \ \vert \  X_1 > \max_{k>1}  X_k$} is
% \begin{equation}\label{eq:SPRT-stat}
$$
    T_{1j} = \left[\frac{\phi(0)}{1-\Phi(\frac{\bareta-x_i}{\barsig})}\right]\left[\frac{\phi(\frac{x_1-\barmu}{\barsig})}{1-\Phi(\frac{\bareta-\barmu}{\barsig})}\right]^{-1}.
$$
% \end{equation}  
If $T_{1j}\geq \frac{1-\beta}{\alpha}$ for all $j$ such that $X_i > X_j$, then verify $\mu_i$ as larger than all $\mu_j$.

\end{theorem}

% See Appendix \ref{apx:SPRT} for the proof. 
We employ Theorem \eqref{thm:LR} to learn the correct top-$K$ rankings (Alg. \ref{alg:sprt-rank}) and set (Alg. \ref{alg:sprt-set}) with probability $1-\alpha$. 
In essence, our algorithms run computation until Procedures 1 and 2 verify the top $K$.
However, the procedures are modified from Section~\ref{sec:goldwasser}: here, they test each null $H_{0ij}$ with the likelihood ratio $T_{ij}$, rather than  p-values $p_{ij}$ \eqref{eq:pval}. 
By SPRT, these tests can be conducted at any $n$ without regard for optional stopping.

\begin{algorithm}
\caption{Rank Stability via SPRT (SPRT-SHAP)}\label{alg:sprt-rank}
\begin{algorithmic}
    \Require Desired rankings $K> 0$, error rate $\alpha \in [0,1]$, total sample budget $n_{max}$, samples between tests $n_{btwn}$%\in \N
    % sample parameters $n_{max}$, $n_{btwn}$, number of features to rank correctly $K$, significance level $\alpha$, 
    \Ensure $\hat\phi_1, \ldots, \hat\phi_d$ whose top-$K$ rankings are correct with probability $\geq 1-\alpha$
    % \State $\hat\phi \gets$ KernelSHAP values with $n_{btwn}$ permutations
    \State $n \gets 0$
    \While{$n < n_{max}$} 
        \State Generate $n_{btwn}$ new samples %of $s_i$ (e.g. local importances)
        \State $n \gets n + n_{btwn}$
        \State $\hat\phi_n \gets$ Feature importances fit on all $n$ samples
        \State $\hat{\Sigma}_n \gets$ Variances of all feature importances
        \State $K' \gets $ Procedure 1 (Alg. \ref{alg:procedure1}) on $\hat\phi_n$ \& $\hat\Sigma_n$, rejecting 
        \Statex \hspace{\algorithmicindent} \qquad\; tests $H_{0ij}$ if $T_{ij}>\frac{1-\beta}{\alpha}$ (Thm. \ref{thm:LR})
        \If {$K'>=K$}
            \State \Return $\hat\phi$, ``Verified''
        \EndIf
    \EndWhile
    \State \Return $\hat\phi$, ``Failed to verify''
\end{algorithmic}
\end{algorithm}

\subsection{Resampling Approach}\label{sec:StableSHAP}

While valid, SPRT is fairly conservative, as it must hold for any number of samples. 
An alternative approach tests with the original, more powerful p-values $p_{ij}$ \eqref{eq:pval}. To account for optional stopping, it throws out all data used to compute $\hat\phi_i$ and $\hat\phi_j$ when null $H_{0ij}$ fails to reject.
As a result, subsequent tests are independent of previous results. 
To guarantee validity, the earlier assumptions of independence, normality, unbiasedness, and oracle variance are still necessary.

This resampling strategy is only more efficient on certain importance scores. 
Shapley Sampling, for example, computes each attribution $\hat\phi_j$ in isolation with separate subsets $S_j$ \eqref{eq:Shapley Sampling}. 
Therefore when $H_{0ij}$ fails to reject at level $\alpha$, only $\hat\phi_i$ and $\hat\phi_j$ must be recomputed; data for all $k\neq i, j$ may be kept. 
This is not the case for KernelSHAP, in which the same set of samples is used to compute all $d$ attributions $\hat\phi_j$.
In that case, removing the entire dataset each time a pairwise test fails to reject is not viable. 

Similarly, resampling may be more efficient than SPRT on global attributions computed in isolation (e.g. average Shapley Sampling, loss-based methods), but not jointly (e.g. average KernelSHAP, surrogate methods). 
Thus, all attributions that may benefit from this approach take the form
$$\hat\phi_j = \frac{1}{n_j} \sum_{i=1}^{n_j} s_j^i.$$
When some pairwise null $H_{0ij}$ fails to reject, the goal is to recompute $\hat\phi_i$ and $\hat\phi_j$ with as few samples as possible for $H_{0ij}$ to subsequently reject. 
Here we present two methods to approximate this.
Let $\tilde{\sigma}_{j}^2$ be the sample variance of $s_j^i$, so \mbox{$\sigma_j^2 = \Var(\hat\phi_j) = \frac{\tilde{\sigma}_j^2}{n_j}$}. 
% Because variance scales at an $O(1/n)$ rate
\citet{goldwasser2025gaussianrankverification} show that when the runner-up has highest p-value, Equation~\eqref{eq:pval} reduces to a Z-test at level $\alpha/2$. Assuming this occurs, 
\begin{equation}\label{eq:Z_test}
    \frac{\hat\phi_i - \hat\phi_j}{\sqrt{\frac{\tilde{\sigma}_i^2}{n_i}+\frac{\tilde{\sigma}_j^2}{n_j}}} < Z_{1-\alpha/2},
\end{equation}
where $Z_{1-\alpha/2}$ is the upper $\frac{\alpha}{2}$ quantile of the standard normal distribution. 

Suppose we want the new number of samples $n_i, n_j$ to be the same $n'$ for both features. Then solving Equation~\eqref{eq:Z_test} for $n'$ leads to the sample size 
\begin{equation}\label{eq:sample size equal}
    n' = \Big(\frac{Z_{1-\alpha/2}}{\hat\phi_i-\hat\phi_j}\Big)^2 (\tilde{\sigma}^2_i+\tilde{\sigma}^2_j).
\end{equation}
Alternatively, we may want the sample size to scale with the variance. Intuitively, more samples should be used to stabilize highly variable features. Defining $n_j' := \frac{\tilde\sigma^2_j}{\tilde\sigma^2_j} n_i'$ and solving for $n_i'$ yields 
\begin{equation}\label{eq:sample size unequal}
    n_i' = 2\Big(\frac{Z_{1-\alpha/2}}{\hat\phi_i-\hat\phi_j}\Big)^2 \tilde\sigma^2_i,\ \  
    n_j' = 2\Big(\frac{Z_{1-\alpha/2}}{\hat\phi_i-\hat\phi_j}\Big)^2 \tilde\sigma^2_j.
\end{equation}
These lower bounds estimate the minimum number of samples needed to obtain an anticipated significant result. To avoid narrowly missing the mark, it is reasonable to choose values of $n'$ that exceed them by a small buffer, e.g. 10\%. It is entirely possible that more optimal choices of $n_i'$ and $n_j'$ exist; we leave this as an open problem.

\begin{algorithm}
\caption{Rank Stability via Resampling (StableSHAP)}\label{alg:StableSHAP}
\begin{algorithmic}
    \Require Desired rankings $K> 0$, error rate $\alpha \in [0,1]$, per-feature sample budget $n_{max}$, intial per-feature samples $n_{init}$, buffer $c\geq 1$
    % sample parameters $n_{max}$, $n_{btwn}$, number of features to rank correctly $K$, significance level $\alpha$, 
    \Ensure $\hat\phi_1, \ldots, \hat\phi_d$ whose top-$K$ rankings are correct with probability $\geq 1-\alpha$
    % \State $\hat\phi \gets$ KernelSHAP values with $n_{btwn}$ permutations
    \State $\hat\phi \gets$ Feature importances fit on $n_{init}$ samples
    \State $\hat{\Sigma} \gets$  Variances of all feature importances
    \While{$n_j < n_{max} \;\forall j$} 
    
        \State $K' \gets $ Procedure 1 (Alg. \ref{alg:procedure1}) on $\hat\phi_n$ \& $\hat\Sigma_n$, rejecting% $H_{0ij}$ if $T_{ij}>\frac{1-\beta}{\alpha}$
        \Statex \hspace{\algorithmicindent} \qquad\; tests $H_{0ij}$ if $T_{ij}>\frac{1-\beta}{\alpha}$ (Thm. \ref{thm:LR})
        \If {$K'>=K$}
            \State \Return $\hat\phi$, ``Verified''
        \Else
            \State $n_i', n_j' \gets $ Est. samples to reject, Eq. \eqref{eq:sample size equal} or \eqref{eq:sample size unequal}.
            \State $n_i', n_j' \gets \min(\lceil c n_i'\rceil, n_{max}), \min(\lceil c n_j'\rceil, n_{max})$
            \State $\hat\phi_i, \hat\phi_j \gets$ Attributions fit on $n_i', n_j'$ samples
            \State $\hat{\Sigma}_i, \hat{\Sigma}_j \gets$  Variances of feature importances
        \EndIf
    \EndWhile
    \State \Return $\hat\phi$, ``Failed to verify''
\end{algorithmic}
\end{algorithm}

Running for the maximal number of samples is not guaranteed to yield $K$ rejections at the desired error tolerance.
Nevertheless, it may suffice even when the anticipated requisite budget is higher, since Equations \eqref{eq:sample size equal} and  \eqref{eq:sample size unequal} rely on plug-in variance estimates.

\section{Experiments}\label{sec:experiments}

% To test the efficacy of our retrospective and top-$K$ methods, we run them on standard estimates of SHAP values. 
% In addition, we evaluated our methods' ability to stabilize rankings from LIME. 
We evaluated our retrospective and top-$K$ methods' ability to stabilize rankings from SHAP and LIME. 
These attributions describe neural network classifiers fit on the Adult Census Income, Portuguese Bank, BRCA, Wisconsin Breast Cancer, and German Credit datasets. Further results in Appendix \ref{apx:extra} demonstrate the efficiency of the resampling approach. 
Appendix \ref{apx:exp} contains more detailed information on these experiments.

\begin{figure}
    \centering
    \includegraphics[width=\columnwidth]{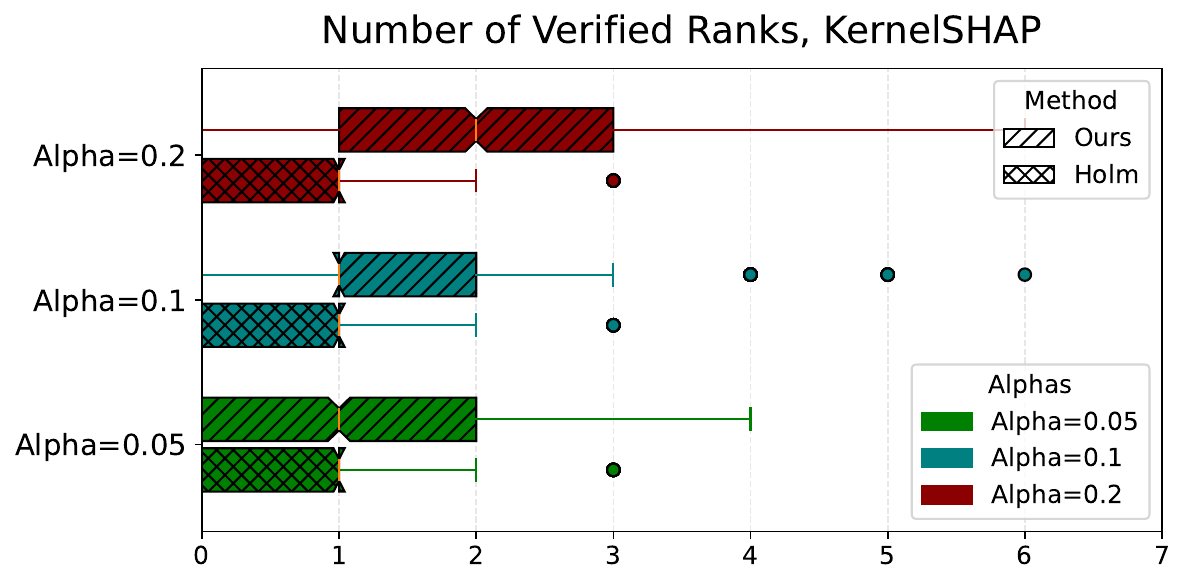}
    \caption{Comparing Number of Stable Features in Retrospect.}
    \label{fig:num_verified}
\end{figure}

\subsection{Retrospective Stability}\label{sec:retroResults}

We tested the efficacy of our retrospective tools %in Section~\ref{sec:retro} 
across a range of datasets, attribution methods, and significance levels. 
To do so, we randomly selected 30 inputs from the test set of each benchmark dataset. 
On each data point, we ran Shapley Sampling and KernelSHAP 50 times. 
For each iteration, we identified the number of stable ranks (Procedure 1) and assessed the stability of the top-$5$ set (Procedure 2).

% We evaluated the validity of the retrospective tools from Section~\ref{sec:retro} across five benchmark datasets, with varying levels of $\alpha$.
% Shapley values for 30 random samples from the test set were estimated 50 times, with both Shapley Sampling and KernelSHAP.
% On these estimates, we identified the number of stable ranks with Procedure 1 (Alg. \ref{alg:procedure1}). 
% We also assessed the stability of the top-$5$ set with Procedure 2 (Alg. \ref{alg:procedure2})

We then computed the family-wise error rate (FWER) of the ranking and set procedures on each input.
The FWER is the fraction of iterations with an error in the ranking or set of supposedly stable features.
Iterations that do not verify any rankings or the top-$K$ set are counted as error-free.
We used the most common top-$K$ ranking as the ground truth; 
this is almost certainly correct, since the Shapley estimators are unbiased.

\begin{table*}
\caption{Maximum error rate of retrospective rank (R) and set (S) procedures, across 30 samples. $K=5$ for set.}
\label{tbl:RetroResults}
\centering
\begin{tabular}{lcc cc|cc|cc|cc|cc|cc}
\toprule
& & & \multicolumn{6}{c}{Shapley Sampling} & \multicolumn{6}{c}{KernelSHAP} \\
\cmidrule(lr){4-9} \cmidrule(lr){10-15}
& & & \multicolumn{2}{c}{$\alpha=0.05$} & \multicolumn{2}{c}{$\alpha=0.1$} & \multicolumn{2}{c}{$\alpha=0.2$} & \multicolumn{2}{c}{$\alpha=0.05$} & \multicolumn{2}{c}{$\alpha=0.1$} & \multicolumn{2}{c}{$\alpha=0.2$} \\
\cmidrule(lr){4-5} \cmidrule(lr){6-7} \cmidrule(lr){8-9} \cmidrule(lr){10-11} \cmidrule(lr){12-13} \cmidrule(lr){14-15}
Dataset & N & D & R & S & R & S & R & S & R & S & R & S & R & S \\
\midrule
\texttt{Adult}  & 32,561 & 12 & 2\%& 4\%& 6\%& 4\%& 10\%& 10\%& 4\%& 4\%& 4\%& 6\%& 8\%& 12\%\\
\texttt{Bank}   & 45,211 & 16 & 2\%& 0\%& 4\%& 2\%& 12\%& 8\%& 2\%& 0\%& 4\%& 0\%& 8\%& 2\%\\
\texttt{BRCA}   & 572    & 20 & 2\%& 2\%& 8\%& 4\%& 10\%& 8\%& 2\%& 2\%& 2\%& 4\%& 10\%& 6\%\\
\texttt{Credit} & 1,000  & 20 & 2\%& 0\%& 8\%& 4\%& 10\%& 6\%& 2\%& 0\%& 2\%& 0\%& 6\%& 2\%\\
\texttt{WBC}    & 569    & 30 & 2\% & 2\%& 8\%& 6\%& 10\%& 8\%& 4\%& 2\%& 6\%& 4\%& 10\%& 10\%\\
\bottomrule
\end{tabular}
\end{table*}

Table \ref{tbl:RetroResults} reports the maximal error rates across the 30 data points. 
% We computed Shapley Sampling and KernelSHAP values on all datasets, then ran our retrospective procedures at multiple levels of $\alpha$.
In all 60 settings, the 30 FWERs are at most $\alpha$, indicating these procedures successfully control the FWER.
This supports the application of \citet{goldwasser2025gaussianrankverification} in the context of feature importances.

Figure \ref{fig:num_verified} shows the number of verified ranks on the German Credit dataset.
We benchmark our method against the approach from \citet{benjamini}, which adjusts one-sided p-values with Holm's method.
This conservative multiple testing procedure has less statistical power than the tests we utilize from \citet{goldwasser2025gaussianrankverification}. 
Experimental results reflect this improvement.
For all error thresholds $\alpha$, our retrospective procedure tends to verify more top ranks than the benchmark. The boxplots corresponding to our method consistently have longer tails, as well as a higher median for $\alpha=0.2$.

Examining our method, the median number of verified ranks rises from one to two as $\alpha$ goes to 0.2. In the most extreme case, the ranking of the top six features is stable. However, a substantial fraction of inputs had only zero or one stable ranking. For $\alpha=0.05$, the single top-ranking feature could not be verified on over 25\% of inputs.
These results emphasize the fragile reliability of Shapley rankings, and the need for algorithms that enforce their stability.

\subsection{Top-K Rank Verification}\label{sec:topKResults}
% \subsubsection{StableSHAP}

\begin{figure}
    \centering
    \includegraphics[width=\columnwidth]{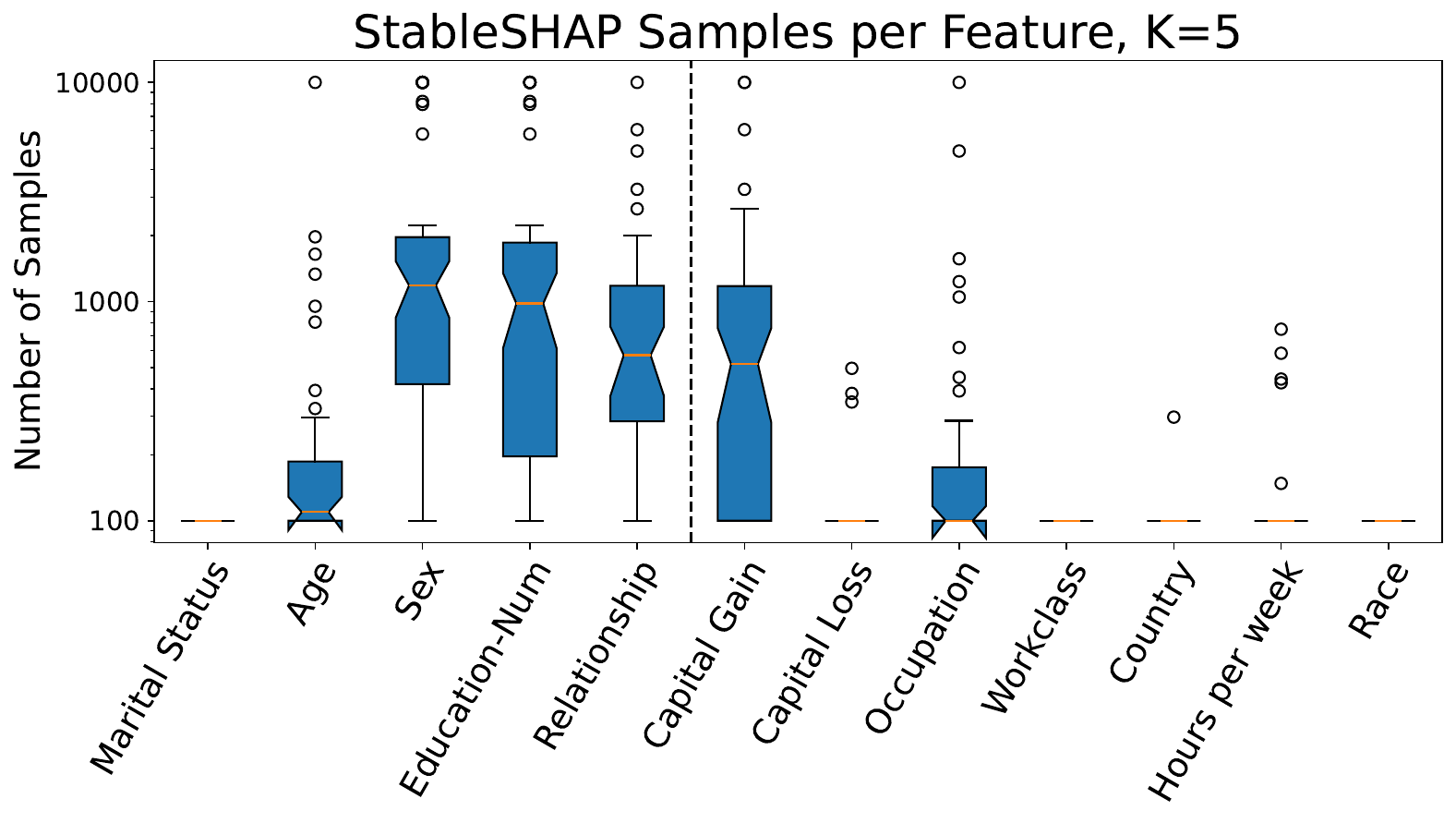}
    % \caption{Number of samples for each feature, ordered by SHAP value. On Adult dataset with $K=5$, $\alpha=0.1$. Variances shown in Figure \ref{fig:StableSHAP_vars}.}
    \caption{StableSHAP Sample Allocation by Feature Ranking.}
    \label{fig:perms_per_feature}
\end{figure}

Empirical results demonstrate that our sampling algorithms identify the $K$ most important features with probability $1-\alpha$. 
Applied to Shapley estimation, we refer to these algorithms as SPRT-SHAP (\ref{alg:sprt-rank}) and StableSHAP (\ref{alg:StableSHAP}). SPRT-SHAP obtains Shapley values with KernelSHAP, whereas StableSHAP is based on Shapley Sampling. 

Table \ref{tbl:TopKResults} summarizes their performance across a range of datasets, desired ranks $K$, and significance levels $\alpha$.
The evaluation metric is the maximum error rate, introduced in Section~\ref{sec:retroResults} for Table \ref{tbl:RetroResults}. As before, FWERs for 30 inputs are computed empirically over 50 runs. 
% As in Section~\ref{sec:retroResults}, it presents the highest FWER over 30 inputs, calculated over 50 runs. 
In all settings for which the algorithms converged within the given sample budget, the FWER was below $\alpha$. 

% Table \ref{tbl:TopKResults} summarizes the performance of our SHAP and LIME methods on a range of datasets, desired ranks $K$, and significance levels $\alpha$. SPRT-SHAP estimated Shapley values using KernelSHAP. On each dataset, we computed FWERs across 100 iterations, then averaged across 10 randomly selected input samples from the test set. In all cases, this empirical FWER was below $\alpha$. 

\begin{table*}
\caption{Maximum error rate of top-$K$ rank (R) and set (S) procedures, across 30 input data points. NAs indicate the procedure does not reject all $K$ tests on the provided inputs.}
%sprtshap_brca_K5 set 0.1; sprtshap_breast_cancer_K5 set 0.1; sprtshap_credit_K5 set 0.2; sprtshap_breast_cancer_K5 set 0.2; StableSHAP_breast_cancer_K2 rank 0.2; StableSHAP_credit_K5 rank 0.1; StableSHAP_breast_cancer_K2 rank 0.1; 
\label{tbl:TopKResults}
\centering
\begin{tabular}{l cc|cc|cc|cc|cc|cc|cc|cc}
\toprule
& \multicolumn{8}{c}{StableSHAP} & \multicolumn{8}{c}{SPRT-SHAP} \\
\cmidrule(lr){2-9} \cmidrule(lr){10-17}
& \multicolumn{4}{c}{\( K=2 \)} & \multicolumn{4}{c}{\( K=5 \)} & \multicolumn{4}{c}{\( K=2 \)} & \multicolumn{4}{c}{\( K=5 \)} \\
\cmidrule(lr){2-5} \cmidrule(lr){6-9} \cmidrule(lr){10-13} \cmidrule(lr){14-17}
& \multicolumn{2}{c}{$\alpha=0.1$} & \multicolumn{2}{c}{$\alpha=0.2$} & \multicolumn{2}{c}{$\alpha=0.1$} & \multicolumn{2}{c}{$\alpha=0.2$} & \multicolumn{2}{c}{$\alpha=0.1$} & \multicolumn{2}{c}{$\alpha=0.2$} & \multicolumn{2}{c}{$\alpha=0.1$} & \multicolumn{2}{c}{$\alpha=0.2$} \\
\cmidrule(lr){2-3} \cmidrule(lr){4-5} \cmidrule(lr){6-7} \cmidrule(lr){8-9} \cmidrule(lr){10-11} \cmidrule(lr){12-13} \cmidrule(lr){14-15} \cmidrule(lr){16-17}
Dataset & R & S & R & S & R & S & R & S & R & S & R & S & R & S & R & S \\
\midrule
\texttt{Adult}  & 8\% & 2\% & 16\% & 16\% & 6\%  & 6\%  & 14\% & 14\% & 0\% & 0\% & 2\% & 0\% & NA & 0\% & NA & 8\% \\
\texttt{Bank}   & 6\% & 2\% & 14\% & 0\%  & 10\% & 10\% & 20\% & 16\% & 0\% & 0\% & 2\% & 2\% & NA & 0\% & NA & 2\% \\
\texttt{BRCA}   & 6\% & 6\% & 14\% & 10\% & 10\% & 10\% & 20\% & 20\% & NA  & 0\% & NA  & 0\% & NA & 0\% & NA & 2\% \\
\texttt{Credit} & 4\% & 2\% & 8\%  & 2\%  & 4\%  & 4\%  & 12\% & 16\% & 0\% & 0\% & 0\% & 0\% & NA & 0\% & NA & 2\% \\
\texttt{WBC}    & 0\%& 4\% & 10\%  & 12\% & 6\%  & 4\%  & 20\%  & 4\% & 0\% & 0\% & 4\% & 0\% & NA& 0\% & 0\%& 0\% \\
\bottomrule
\end{tabular}
\end{table*}

StableSHAP is highly adaptive to the significance level. 
Its empirical FWERs were closest to $\alpha$, getting up to a 10\% FWER with $\alpha=0.1$ and 20\% with $\alpha=0.2$.
In contrast, SPRT-SHAP produced more stable top-$K$ rankings, if it converged at all.
This matches our intuition that SPRT is more conservative, requiring a higher evidence threshold to reject.

Moreover, StableSHAP allocates computation in a highly efficient manner. %samples subsets $S_j^i$ 
% Figure \ref{fig:perms_per_feature} demonstrates that it adaptively samples subsets $S_j^i$ on the features whose rankings are both ambiguous and relatively high, especially if they are high-variance. 
It adaptively focuses on the features whose rankings are both ambiguous and relatively high.
Figure \ref{fig:perms_per_feature} displays its sample allocations for a given input from the Adult dataset, running 50 times at $\alpha=0.1$.
% Across the 50 runs for a single input, Figure \ref{fig:perms_per_feature} displays its number of samples for each feature, ordered by SHAP value.
StableSHAP uses only the initial 100 samples for the highest-ranked feature, Marital Status, as it wins by a wide margin.
It also avoids precise estimation of ranks beyond the top $K$ and its runner-up, Capital Gain.
% For ranks beyond the top $K$ and its runner-up, it also tends to use only the initial 100 samples.
The main exception, Occupation, has considerably higher variance, thereby raising its p-values (Figure \ref{fig:StableSHAP_vars}).

% % To rank the top $K$ features, extra samples are only regularly needed for the top $K+3$. 
% It also uses only the initial 100 samples for most ranks beyond the top $K$.
% The only exceptions are the runner-up $K+1$, as well as rank $K+3$, which has considerably higher variance (Figure \ref{fig:StableSHAP_vars}).
% % The only features beyond the top $K$ that typically require more than the initial 100 samples are ranks $K+1$ and $K+3$. 

In contrast, most existing Shapley algorithms allot the same budget for all features. 
Figure \ref{fig:boxplots} in Appendix \ref{apx:extra} compares the performance of StableSHAP and Shapley Sampling, given the same inputs and computational budget. 
By and large, StableSHAP demonstrates improved stability, with fewer misranked features. 

\begin{figure}
    \centering
    \includegraphics[width=\columnwidth]{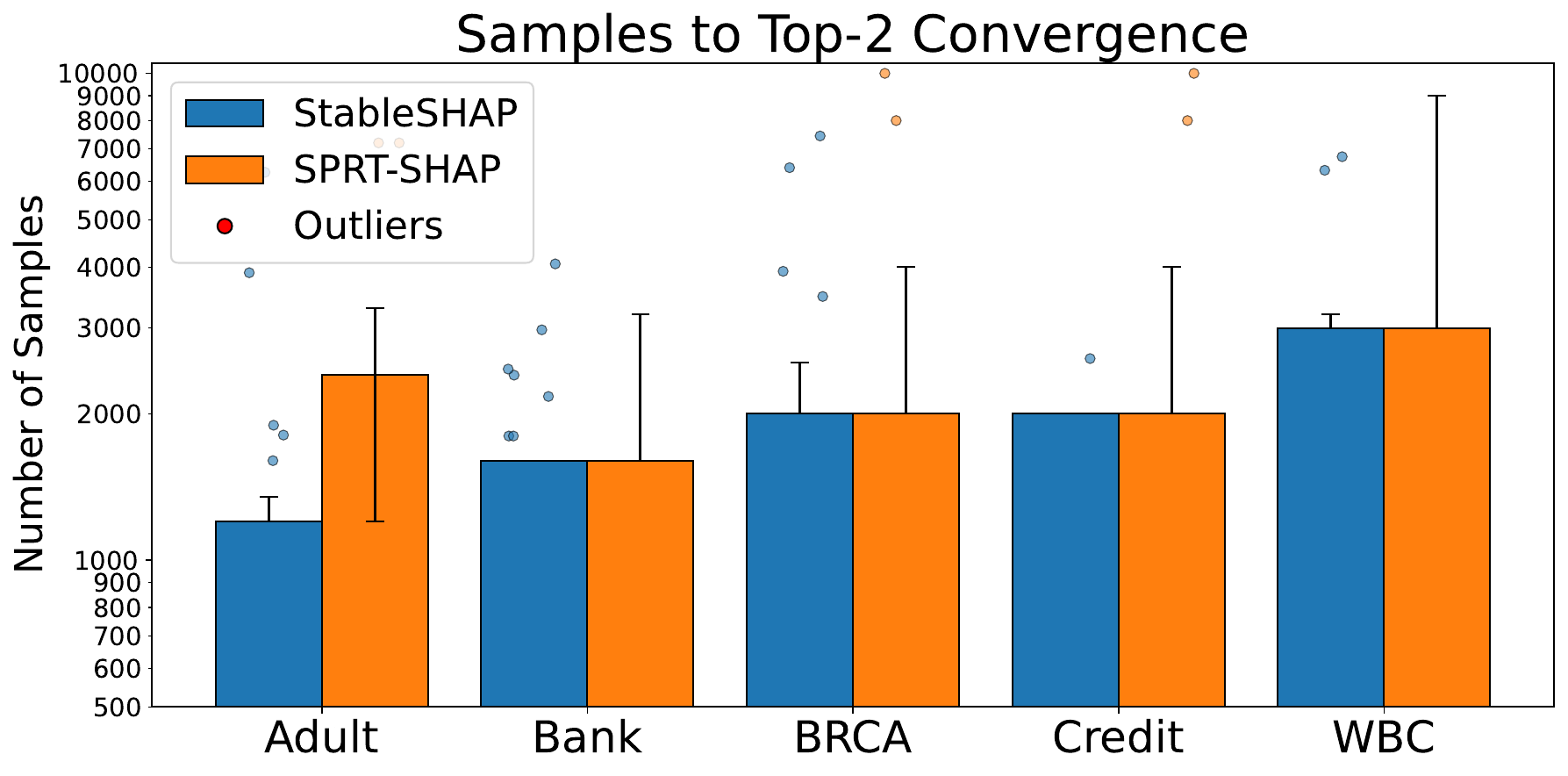}
    \caption{Sampling Efficiency of StableSHAP vs SPRT-SHAP.}
    \label{fig:shap_n_samples}
\end{figure}

Figure \ref{fig:shap_n_samples} visualizes the average runtime of StableSHAP and SPRT-SHAP. 
The methods share the same minimal and maximal budget, and explain the same data. 
While both often achieve top-$2$ rank stability with the initial number of samples, StableSHAP is generally more efficient. 
SPRT-SHAP often takes a larger number of samples to converge, as evidenced by its higher third quartiles. 
% On three of the datasets, they both require about 2000 samples, whereas SPRT-SHAP is considerably faster on the other two. 
% Granted, SPRT-SHAP typically failed to converge for larger values of $K$. 
Overall, these sample budgets are similar to the defaults in the \texttt{shap} package.% (Appendix \ref{apx:expSHAP}). 
% Moreover, they are much faster than computing the SHAP values in closed-form. 
% Computing each SHAP value on BRCA, for example, requires $\mathcal{O}(2^{30}\approx 10^9)$ samples. 

\subsection{LIME Experiments}%\label{apx:LIME Control}

\begin{table*}
\caption{Maximum error rate (\%) for top-$5$ ranking procedure on LIME with $K$-Lasso.}
\label{tbl:LIMEResults}
\centering
\begin{tabular}{lcccc}
\toprule
\multirow{2}{*}{Dataset} & \multicolumn{1}{c}{$K=2$} & \multicolumn{1}{c}{$K=2$} & \multicolumn{1}{c}{$K=5$} & \multicolumn{1}{c}{$K=5$} \\
                         & \multicolumn{1}{c}{$\alpha=0.1$} & \multicolumn{1}{c}{$\alpha=0.2$} & \multicolumn{1}{c}{$\alpha=0.1$} & \multicolumn{1}{c}{$\alpha=0.2$} \\
\midrule
\texttt{Adult}  & 0\% & 2\%& NA& NA \\
\texttt{Bank}   & 2\% & 0\% & 0\% & 0\% \\
\texttt{BRCA}   & 2\%& 0\%& NA& NA \\
\texttt{Credit} & 6\% & 8\%& 0\% & 0\% \\
\texttt{WBC}    & 2\%& 2\%& 0\% & 0\% \\
\bottomrule
\end{tabular}
\end{table*}

% \subsubsection{Methodology}\label{apx:S-LIME}

LIME, introduced in Section \ref{sec:LIME}, fits an interpretable model on data randomly generated around a point of interest.
The default model in the \texttt{lime} Python package is $K$-Lasso, which iteratively selects $K$ features along the Least-Angle Regression (LARS) path. Our experiments modify Algorithm \ref{alg:StableSHAP} to achieve top-$K$ rank stability for LIME with $K$-Lasso.
% We modify the resampling algorithm (Alg. \ref{alg:StableSHAP}) for the context of LIME with $K$-Lasso.

LIME regenerates data at each step along the LARS path, so our methods are not applicable in their current form. 
Rather, to ensure the FWER is controlled at level-$\alpha$, a multiple testing correction must be used.
We apply the Bonferroni correction, verifying each feature selection at level $\alpha/K$.
Note this assumes LARS does not deselct any features, in which case more than $K$ tests would be performed.

At each step, LARS iteratively chooses the predictor that is most correlated with the current residuals.
These scaled correlations $\hat c_j$ are asymptotically normal by the central limit theorem. 
Therefore top-$K$ rank stability can be achieved by running either algorithm in Section~\ref{sec:algorithms} at level $\alpha/K$ until the top rank is stable, for all $K$ steps. 

We did not implement our algorithms on LIME from scratch. 
Rather, we repurposed an existing method, S-LIME, to adhere generally within our framework \citep{S-LIME}. More details are in Appendix \ref{apx:lime}

% \subsubsection{Results}

We ran the adjusted S-LIME procedure on 5 datasets at \mbox{$\alpha=0.1$} and $0.2$.
As before, we compare 30 inputs, computing the error rate with 50 runs across each.
Table \ref{tbl:LIMEResults} displays the experimental results.

For all inputs, the error rate is always controlled at level $\alpha$. 
In fact, this procedure is very conservative, with a maximal error rate of 2\% for four out of five datasets.
This may be attributed to its use of the Bonferroni correction, lowering the significance threshold by a factor of 5.
Future experiments could run our top-$K$ algorithms on LIME with OLS, requiring no such correction.

\section{Discussion}\label{sec:discussion}

In this paper we present methods to obtain stable orderings of feature importance scores. 
For a user-defined error rate $\alpha$, 
our retrospective procedures verify the highest observed rankings and top-$K$ set.
Our top-$K$ algorithms efficiently run computation until the population rankings are attained with high probability.
% Our algorithms guarantee that the probability of an incorrect Shapley ranking or LIME selection of a top-$K$ feature is at most $\alpha$. 
% While the statistical guarantee relies on asymptotic arguments, a sufficiently large number of samples is always used in standard practice.
Our statistical guarantees, contingent on normal assumptions or asymptotic arguments, are unanimously justified by empirical results. 
%(Appendix \ref{apx:exp}).

% StableSHAP can be used for other applications, ML and non ML.
Our methods can be used to rank Shapley values in any context, not specifically for feature attributions. 
% For example, it can be used for Shapley feature importances that sample from conditional distributions  \citep{aas, frye}. 
Other use cases include feature selection  \citep{cohen07}, federated learning  \citep{GTG}, data valuation  \citep{valuation}, multi-agent RL  \citep{multi-agent}, and ensembling  \citep{ensembling}.
Outside of ML, they have been applied
% Further, Shapley values have numerous applications outside of ML. They were originally studied in game theory, allocating credit to players in a cooperative game. Works have applied them 
in fields as diverse as ecology \citep{ecology}, online advertising \citep{ads}, supply chain management \citep{supply_chain}, and financial portfolio optimization \citep{finance}. 

In addition to their analytical utility, our top-$K$ methods run relatively efficiently. 
The resampling approach adaptively budgets computation towards the more important features. 
SPRT for Shapley importances enables use of KernelSHAP, a highly efficient algorithm.
% SPRT-SHAP enables use of more efficient algorithms, specifically KernelSHAP. 
% However, it is fairly conservative by nature, in order to reject at any time. 
While SPRT is fairly conservative due to its anytime-valid requirement, future work could use more relaxed procedures that restrict the number of potential rejection times.
Testing procedures that do so have been proposed in the context of clinical trials, e.g. \citet{pocock,O'Brian,peto1976design}.

Our methods provide concrete statistical guarantees on their rankings and selections, with higher power and more reasonable assumptions than prior work \citep{shap@k, topKbandit, S-LIME}. 
Moreover, they may be applied in conjunction with methods that stabilize the attributions themselves \citep{controlSHAP, mitchell_perms}.

\bibliography{refs}

 %%%%%%%%%%%%%%%% APPENDIX %%%%%%%%%%%%%%%%
\newpage
\appendix
\onecolumn

\section{Procedures 1 and 2}

\begin{algorithm}
\caption{Procedure 1 \citep{goldwasser2025gaussianrankverification}}\label{alg:procedure1}
\begin{algorithmic}
    \Require Significance level $\alpha$, ordered data $X_1 > X_2 > \ldots > X_d$
    \Ensure Integer $K\geq 0$ verifying the top-$K$ population rankings are correct
 with probability at least $1-\alpha$
    \State $K \gets 0$ \Comment{Number of verified ranks}
    \While{$K < d-1$} 
        \State $i \gets K+1$ \Comment{Index of tested feature}
        \For{$j \in [i+1:d]$}
                    \State Test $H_{0ij}:\{\mu_i < \mu_j \ \vert \  X_i>\max_{k>i} X_k\}.$ \Comment{Use original p-value $p_{ij}$ or SPRT statistic $T_{ij}$}
            % \State $p_{ij} \gets $ p-value from Eq. \eqref{eq:pval} \Comment{Testing $H_{0ij}:\{\mu_i < \mu_j \vert X_i>\max_{k>i} X_k\}.$}
        \EndFor
        % \If{all nulls $H_{0ij}$ reject} \Comment{Original procedure: Test $H_{0i}$ with  $p_i^* \gets \max_j p_{ij}$}
        \If{all nulls $H_{0ij}$ reject} \Comment{Equivalently, test $H_{0i}$ with $p_i^* \gets \max_j p_{ij}$ or $T_{i*}^n \gets \min_j T_{ij}$ (SPRT)}
            \State $K \gets K+1$
            % \State $i \gets i+1$
        \Else
            \State \Return $K$
        \EndIf
    \EndWhile
    \If{$K=d-1$}
        \State $K \gets d$
    \EndIf
    \State \Return $d$
\end{algorithmic}
\end{algorithm}

\begin{algorithm}
\caption{Procedure 2 \citep{goldwasser2025gaussianrankverification}}\label{alg:procedure2}
\begin{algorithmic}
    \Require Significance level $\alpha$, ordered data $X_1 > X_2 > \ldots > X_d$
    \Ensure Boolean whether top-$K$ population set is correct 
    % rankings $\mu_1 >\ldots > \mu_K > \max_{j>K} \mu_j$ 
    with probability at least $1-\alpha$
    \For{$i \in [1:K]$} 
        \For{$j \in [K+1:d]$}
                            \State Test $H_{0ij}:\{\mu_i < \mu_j \ \vert \  X_i>\max_{k>K} X_k\}.$  \Comment{Use original p-value $p_{ij}$ or SPRT statistic $T_{ij}$}
            % \State $p_{ij} \gets $ p-value from Eq. \eqref{eq:pval} \Comment{Testing $H_{0ij}:\mu_i < \mu_j \ \vert \  X_i>\max_{k>K} X_k.$}
        \EndFor
        % \State $p_i^* \gets \max_j p_{ij}$
        % \If{$p_i^* > \alpha$}
                \If{some null $H_{0ij}$ fails to reject}  \Comment{i.e. Test $H_{0i}$ with $p_i^* \gets \max_j p_{ij}$ or $T_{i*}^n \gets \min_j T_{ij}$ (SPRT)}
            \State \Return False \Comment{Cannot verify i\textsuperscript{th} rank belongs in top-$K$ set}
        \EndIf
    \EndFor
    \State \Return True
\end{algorithmic}
\end{algorithm}

\section{Top-$K$ Set Algorithms}\label{apx:set}

Algorithms \ref{alg:sprt-set} and \ref{alg:rank-set} stabilize the top $K$ set. Like Algorithms \ref{alg:sprt-rank} and \ref{alg:StableSHAP}, they employ the SPRT and resampling approaches, respectively.

\begin{algorithm}
\caption{Set Stability via SPRT}\label{alg:sprt-set}
\begin{algorithmic}
    \Require Desired set size $K>0$, error rate $\alpha \in [0,1]$, total sample budget $n_{max}$, number of samples between tests $n_{btwn}$, samples between tests $n_{btwn}$. 
    \Ensure Estimated feature importances $\hat\phi_1, \ldots, \hat\phi_d$ whose top-$K$ set is correct with probability $\geq 1-\alpha$.
    \State $n \gets 0$
    \While{$n < n_{max}$} 
        \State Generate $n_{btwn}$ new samples %of $s_i$ (e.g. local importances)
        \State $n \gets n + n_{btwn}$
        \State $\hat\phi_n \gets$ Feature importances fit on all $n$ samples
        \State $\hat{\Sigma}_n \gets$ Variances of all feature importances
        \State Status $\gets $ Procedure 2 (Alg. \ref{alg:procedure2}) on $\hat\phi_n$ \& $\hat\Sigma_n$, rejecting $H_{0ij}$ if $T_{ij}>\frac{1-\beta}{\alpha}$
        \If {Status$==$``Verified''}
            \State \Return $\hat\phi$, ``Verified''
        \EndIf
    \EndWhile
    \State \Return $\hat\phi$, ``Failed to verify''
\end{algorithmic}
\end{algorithm}

\begin{algorithm}
\caption{Set Stability via Resampling}\label{alg:rank-set}
\begin{algorithmic}
    \Require Desired rankings $K> 0$, error rate $\alpha \in [0,1]$, per-feature sample budget $n_{max}$, intial per-feature samples $n_{init}$, buffer $c\geq 1$.
    \Ensure Estimated feature importances $\hat\phi_1, \ldots, \hat\phi_d$ whose top-$K$ rankings are correct with probability $\geq 1-\alpha$.
    % \State $\hat\phi \gets$ KernelSHAP values with $n_{btwn}$ permutations
    \State $\hat\phi \gets$ Feature importances fit on $n_{init}$ samples
    \State $\hat{\Sigma} \gets$  Variances of all feature importances
    \While{$n_j < n_{max} \;\forall j$} 
        \State Status $\gets $ Procedure 2 (Alg. \ref{alg:procedure2}) on $\hat\phi_n$ \& $\hat\Sigma_n$, rejecting $H_{0ij}$ if $T_{ij}>\frac{1-\beta}{\alpha}$
        \If {Status$==$``Verified''}
            \State \Return $\hat\phi$, ``Verified''
        \Else
            \State $n_i', n_j' \gets $ Estimated samples to reject, Eq. \eqref{eq:sample size equal} or \eqref{eq:sample size unequal}.
            \State $n_i', n_j' \gets \min(\lceil c n_i'\rceil, n_{max}), \min(\lceil c n_j'\rceil, n_{max})$
            \State $\hat\phi_i, \hat\phi_j \gets$ Attributions fit on $n_i', n_j'$ samples
            \State $\hat{\Sigma}_i, \hat{\Sigma}_j \gets$  Variances of feature importances
        \EndIf
    \EndWhile
    \State \Return $\hat\phi$, ``Failed to verify''
\end{algorithmic}
\end{algorithm}

% \subsection{Results}

% Tables \ref{tbl:RetroResults} and \ref{tbl:TopResults} demonstrate the validity of the retrospective and top-$K$ set methods.

\section{SPRT Proof}\label{apx:SPRT-proof}

Here we prove Theorem \ref{thm:LR}, presenting the likelihood ratio necessary for SPRT. 
Returning to the terminology of Section~\ref{sec:goldwasser}, let $X_j\sim\cN(\mu_j, \sigma_j^2)$ with independence and known $\sigma_j$; sort \mbox{$X_1 > X_2 > \ldots X_d$}.

Our objective is to verify feature $i$ as having the largest mean in a set of random variables, given it has the largest observed value $X_i$.
% larger mean than features $j$, given that $X_i$ is larger than all the variables we compare it to. 
Given a valid test to do so, Procedure 1 iterates from $i=1$ onwards until a failure to reject at $i=K+1$. At each step, $i$ is tested against all lower ranks. 
Procedure 2 conducts this test on $i=1$ through $K$ against features $j=K+1, \ldots, d$. 
Because rankings within the top-$K$ set do not matter, it only conditions on $\{X_1 > \max_{j>K}X_j\}$.

Without loss of generality, let $i=1$. Define the null hypothesis
\begin{align*}
    H_{0}:\ &\mu_1 \text{ \textit{not} best} \ \vert \  X_1 \text{ wins} \\
    \Longleftrightarrow\qquad &\mu_1 \leq \max_{j>1} \mu_j \ \vert \  X_1 > \max_{k>1} X_k\\
    \Longleftrightarrow\qquad &\bigcup_{j>1} \underbrace{\{\mu_1 \leq \mu_j \ \vert \  X_1 > \max_{k>1} X_k \}}_{H_{0j}}.
\end{align*}
Analogously define the alternate hypothesis
\begin{align*}
    H_{1}:\ &\mu_1 \text{ \textit{is} best} \ \vert \  X_1 \text{ wins} \\
    \Longleftrightarrow\qquad &\mu_1 > \max_{j>1} \mu_j \ \vert \  X_1 > \max_{k>1} X_k\\
    \Longleftrightarrow\qquad &\bigcap_{j>1} \underbrace{\{\mu_1 > \mu_j \ \vert \  X_1 > \max_{k>1} X_k \}}_{H_{1j}}.
\end{align*}
A classical result states that a valid p-value for a union null hypothesis is the maximum p-value of its constituent nulls \citep{Berger1982}. 
Therefore it suffices to construct valid tests for all $H_{0j}$; rejecting $H_0$ when all tests reject is a valid level-$\alpha$ procedure.

Ostensibly, SPRT tests $H_{0j}$ with a likelihood ratio $T_{1j}$, not a p-value. 
% The null is rejected when $T \geq \gamma_1 = \frac{1-\beta}{\alpha}$.
However, $T_{1j}$ may be interpreted probabilistically: Because SPRT is a valid test, the probability under the null that $T$ exceeds threshold $\gamma_1 = \frac{1-\beta}{\alpha}$ is at most $\alpha$. 
Therefore the SPRT p-value is conceivably the probability of randomly obtaining a higher $T$ than the observed quantity, given that the null is true.
Amongst all nulls $H_{0j}$, the p-value is highest for the lowest $T_{1j}$. 
So it suffices to compute $T_{1j}\; \forall j>1$, and reject $H_0$ if the smallest $T_{1j} \geq \gamma_1$. (Equivalently, $T_{1j}$ must exceed $\gamma_1$ for all $j$.)

Accepting the null is not a concern for reasonable $\alpha$ and $\beta$. 
For example, when $\alpha=0.05$ and $\beta=0.2$, the null is accepted when $T_{1j}$ is below roughly $0.21$. 
In practice, however, $T_{1j}$ should never even be below 1. Conditioning on the selection event that $X_1 > X_j$, the data will always be likelier when the means mirror this discrepancy with $\Delta>0$.

Let $\Delta = \mu_1-\mu_j$. Under the null $\Delta\leq 0$, and under the alternate $\Delta > 0$.
Also let $A_1 = \{X_1 > \max_{k>1} X_k\}$, the event that $X_1$ wins.
The $j$\textsuperscript{th} likelihood ratio is then

\begin{equation*}
    T_{1j} = \frac{\max_{\Delta > 0} \Prob_{\mu_1-\mu_j=\Delta}(X_1, \ldots, X_d \ \vert \  A_1)}{\max_{\Delta \leq 0} \Prob_{\mu_1-\mu_j=\Delta}(X_1, \ldots, X_d \ \vert \  A_1)}.
\end{equation*}

Following the same argument of \citet{goldwasser2025gaussianrankverification}, these probabilities may be simplified by conditioning on additional variables. 
The ensuing test will be valid at level-$\alpha$ for all possible realizations; as a result, the unconditional test would still be valid after marginalizing them out.

In particular, we condition on the values of non-tested variables $X_{-1j}$.
We also condition on \mbox{$U(X) = \frac{X_1}{\sigma_1^2} + \frac{X_j}{\sigma_j^2}$} taking its realized value, \mbox{$u=\frac{x_1}{\sigma_1^2}+\frac{x_j}{\sigma_j^2}$}. 
The purpose of this is to remove the influence of a nuisance parameter, which would otherwise prohibit inference on $\Delta$. 
We refer the reader to the manuscript for greater detail.
The ratio that is tested instead is

\begin{equation}\label{eq:SPRT-distr}
    T_{1j} = \frac{\max_{\Delta > 0} \Prob_{\mu_1-\mu_j=\Delta}(X_1 \ \vert \  A_1, X_{-1j}, U)}{\max_{\Delta \leq 0} \Prob_{\mu_1-\mu_j=\Delta}(X_1\ \vert \  A_1, X_{-1j}, U)}.
\end{equation}

Ignoring constant factors, the conditional likelihood is proportional to the following:
\begin{equation}\label{eq:cond-likelihood}
    X_{1}\ \vert \  \{A_1, X_{k\neq 1, j}, U\} \propto \exp\left[- \left(\frac{1}{2\sigma_1^2} + \frac{\sigma_j^2}{2\sigma_1^4}\right)X_1^2 + \left(\frac{\sigma_j^2 u +\Delta}{\sigma_1^2}\right)X_1\right]\textbf{1}_{A_1}.
\end{equation}
\citet{goldwasser2025gaussianrankverification} showed that this likelihood under the null is maximized at $\Delta=0$.
The conditional distribution for $X_1$ that results is a truncated normal. Its parameters are defined in Section~\ref{sec:goldwasser}: Mean $\barmu = \frac{\sigma_j^2 x_1 + \sigma_1^2 x_j}{\sigma_1^2 + \sigma_j^2}$, variance $\barsig^2 = \frac{\sigma_1^4}{\sigma_1^2+\sigma_j^2}$, and truncation at $\bareta=\max(\barmu, \max_{k\neq 1,j} x_k)$. 
The denominator of \eqref{eq:SPRT-distr} is thus
\begin{equation}\label{eq:SPRT-denom}
    \frac{\phi(\frac{x_1-\barmu}{\barsig})\frac{1}{\barsig}}{1-\Phi(\frac{\bareta-\barmu}{\barsig})}.
\end{equation}
For arbitrary $\Delta$, the conditional likelihood is also a truncated normal. Its proportional density \eqref{eq:cond-likelihood} may be rearranged to complete the square.
\begin{align*}%\label{eq:complete_square}
    X_{1}\ \vert \  \{A_1, X_{k\neq 1, j}, U\} &\propto \exp\left[ \left(-\frac{\sigma_1^2+\sigma_j^2}{2\sigma_1^4}\right)X_1^2 + \left(\frac{\sigma_j^2 u + \Delta }{\sigma_1^2}\right)X_1\right]\textbf{1}_{A_1}\nonumber\\
    &\propto \exp\left[-\frac{\sigma_1^2+\sigma_j^2}{2\sigma_1^4}\left(X_1^2 - \frac{2\sigma_1^2(\sigma_j^2 u +\Delta)}{\sigma_1^2+\sigma_j^2} X_1\right) \right]\textbf{1}_{A_1}\nonumber\\
    &\propto \exp\left[-\frac{\sigma_1^2+\sigma_j^2}{2\sigma_1^4}\left(X_1 - \frac{\sigma_1^2\sigma_j^2 u + \sigma_1^2\Delta}{\sigma_1^2+\sigma_j^2}\right)^2\right]\textbf{1}_{A_1},
\end{align*}
This is a truncated normal distribution again with variance $\barsig^2$ and truncated at $\bareta$. Its mean decomposes to the following:

\begin{equation*}
    \tilde{\mu}_{1j}^\Delta = \frac{\sigma_1^2\sigma_j^2 u + \sigma_1^2\Delta}{\sigma_1^2+\sigma_j^2}= \frac{\sigma_j^2 x_1 + \sigma_1^2 x_j + \sigma_1^2\Delta}{\sigma_1^2+\sigma_j^2}.
\end{equation*}
The numerator of Equation~\eqref{eq:SPRT-distr} is 
\begin{equation}\label{eq:SPRT-num}
    \max_{\Delta > 0} \frac{\phi(\frac{x_i-\tilde{\mu}_{1j}^\Delta}{\barsig})\frac{1}{\barsig}}{1-\Phi(\frac{\bareta-\tilde{\mu}_{1j}^\Delta}{\barsig})}.
\end{equation}

This expression cannot be manipulated into a closed-form solution for the optimal $\Delta$, due to the integral in the denominator. 
It could be optimized numerically, but doing so would be relatively slow. 
Furthermore, the SPRT algorithm must compute $T_{ij}$ for many $i,j$, and $n$, so this is impractical. 
Instead, a practical solution optimizes only the numerator of \eqref{eq:SPRT-num}. 
The resulting choice of $\Delta$ may be slightly sub-optimal, producing a test that is barely more conservative. 
\begin{align*}
    \Delta^* &= \argmax_{\Delta>0} \phi(\frac{x_i-\tilde{\mu}_{1j}^\Delta}{\barsig}) \\
    &= \argmax_{\Delta>0} \exp\left[-\frac{(x_1-\tilde{\mu}_{1j}^\Delta)^2}{2\barsig^2}\right]\\
    &= \argmin_{\Delta > 0} (x_1 - \tilde{\mu}_{1j}^\Delta)^2\\
    &= \argmin_{\Delta > 0} \left(\frac{\sigma_1^2 x_1 + \sigma_j^2 x_1}{\sigma_1^2+\sigma_j^2}-\frac{\sigma_j^2 x_1 + \sigma_1^2 x_j + \sigma_1^2\Delta}{\sigma_1^2+\sigma_j^2}\right)^2\\
    &= \argmin_{\Delta > 0} \left(\frac{\sigma_1^2}{\sigma_1^2+\sigma_j^2}(x_1+x_j-\Delta)\right)^2\\
    &= x_1-x_j.
\end{align*}

This is an intuitive choice: The optimal difference in means $\Delta^*$ is the observed difference, $x_1 - x_j$.
% Moreover, at $\Delta^* = x_1-x_j$,
Furthermore, the conditional mean of $X_1$ under the alternate is the observed value $x_1$:
$$    \tilde{\mu}_{1j}^{\Delta^*} = \frac{\sigma_j^2 x_1 + \sigma_1^2 x_j + \sigma_1^2 x_1-\sigma_1^2 x_j}{\sigma_1^2+\sigma_j^2} = \frac{\sigma_j^2 x_1 + \sigma_1^2 x_1}{\sigma_1^2+\sigma_j^2} = x_1.
$$
Therefore the alternate likelihood \eqref{eq:SPRT-num} is
\begin{equation*}
    \frac{\phi(0)\frac{1}{\barsig}}{1-\Phi(\frac{\bareta-x_1}{\barsig})}.
\end{equation*}
Composing this with \eqref{eq:SPRT-denom} yields
$$T_{1j} = \left[\frac{\phi(0)}{1-\Phi(\frac{\bareta-x_1}{\barsig})}\right]\left[\frac{\phi(\frac{x_i-\barmu}{\barsig})}{1-\Phi(\frac{\bareta-\barmu}{\barsig})}\right]^{-1}.$$
Finally, the test statistic for $H_0$ is $\min_{j>1} T_{1j}$.

\section{Correlated Testing}\label{apx:correlation}

The methodology we have presented assumes estimated ranks are independent. 
This assumption is necessary in order to apply the hypothesis tests in Equation~\eqref{eq:pval} and Theorem \ref{thm:LR}.
As described in Section~\ref{sec:retro}, however, feature importances may be weakly correlated with one another. 
Violating this assumption is a limitation of our methodology.
Fortunately, this limitation does not seem to affect practice, as the empirical FWERs of KernelSHAP are all controlled at level $\alpha$.

Nevertheless, it turns out that alternative tests can be constructed which take correlation into account. 
Unfortunately, they require making assumptions that are at least as concerning as indepedence itself. 

To test each hypothesis $H_{0ij}$ \eqref{eq:nullH0i}, it is necessary to construct a density that isolates the parameter of interest, $\mu_i-\mu_j$. 
The joint density of all observations $X_{1:d}$ has a host of ``nuisance parameters'' --- other unknown quantities that prohibit inference on $\mu_i-\mu_j$. 
To remove their inference, a requisite step conditions on the non-tested variables $X_k=x_k$, where $k\neq i, j$. 
% To test each hypothesis $H_{0ij}$ \eqref{eq:nullH0i}, it is necessary to condition on the non-tested variables $X_k=x_k$, where $k\neq i, j$. 
% This is an important step in reducing the density to the parameter upon which we perform inference, $\mu_i-\mu_j$. 
When all variables are independent, the nuisance parameters $\mu_k$ disappear in the density of $X_{i,j}\vert X_{k\neq i, j}$. 
The remaining density can be rewritten in terms of $\mu_i-\mu_j$ and a final nuisance parameter, $\mu_j$.
% Conditioning on $U(X)=\frac{X_i}{\sigma_i^2}+\frac{X_j}{\sigma_j^2}=u$ with the observed value eliminates this nuisance parameter.
% Refer to \citet{goldwasser2025gaussianrankverification} for the complete argument, similar to the proof in Appendix~\ref{apx:SPRT-proof}.

When variables are correlated, the conditional distribution $X_{i,j}\vert X_{k\neq i, j}$ does \textit{not} eliminate the nuisance parameters $\mu_{k\neq i, j}$. 
This is because the conditional means $\mathbb{E}[X_{i,j}\ \vert \  X_{k\neq i, j}]$ depend on $\mu_k$.
To see this, let $X\sim\cN(\mu, \Sigma)$, with indices $a=\{i,j\}$ and $b=[d]\backslash\{i,j\}$. 
Express the unconditional mean and variance as
\begin{equation*}
    \mu = \begin{bmatrix}
        \mu_a\\
        \mu_b
    \end{bmatrix}
    \text{\quad and \quad}
    \Sigma = \begin{bmatrix}
        \Sigma_{aa} & \Sigma_{ab}\\
        \Sigma_{ba} & \Sigma_{bb}
    \end{bmatrix}.
\end{equation*}
Classical results state the conditional distribution is normal with means and variance
\begin{align*}
    \mu_{a|b}&= \mu_a + \Sigma_{ab} \Sigma_{bb}^{-1} (x_b - \mu_b),\\
\Sigma_{a|b} &= \Sigma_{aa} - \Sigma_{ab} \Sigma_{bb}^{-1} \Sigma_{ba}
\end{align*}
Thus, the density of $X_{i,j}\vert X_{k\neq i, j}$ still contains all means $\mu_{k\neq i, j}$, vectorized in $\mu_b$. 
These nuisance parameters prohibit inference on $\mu_i-\mu_j$. 

A strong assumption enables us to circumvent this restriction. 
The conditional means $\mu_i$ have their own ranking: Conditioned on $X_k$, we may compare $\mu_{i\vert b}$ and $\mu_{j\vert b}$. 
For all $i$ and $j$, we may assume the unconditional ordering of $\mu_i$ and $\mu_j$ is equal to the conditional ordering of $\mu_{i\vert b}$ and $\mu_{j\vert b}$. 
When this holds, a valid test of $H_{0ij}$ performs inference on the null that $\mu_{i\vert b}-\mu_{j\vert b}$.

It is possible to construct a test on these conditional means that takes their correlation $\rho_{ij}$ into account, using the conditional variance $\Sigma_{a\vert b}$. The argument to do so is identical to that of \citet{goldwasser2025gaussianrankverification}:
\begin{enumerate}
    \item Express the density in terms of the difference in means $\mu_{i\vert b}-\mu_{j\vert b}$ and a single nuisance parameter.
    \item Conditioning on the ancillary statistic removes the nuisance parameter, and expresses the distribution in terms of $X_i$.
    \item Apply the null hypothesis, under which $\mu_{i\vert b}-\mu_{j\vert b}=0$. 
    \item Completing the square reveals the mean and variance of a truncated normal distribution.
    \item Identify the truncation event for $X_i$ winning in terms of $X_{k\neq i, j}$ and the ancillary statistic.
\end{enumerate}

This approach may be followed for top-$K$ SPRT algorithms. The optimal densities under the null and alternative can be calculated, producing a likelihood ratio akin to Theorem \eqref{thm:LR}.

We do not include these results and their proofs because the assumption of mean order preservation is very strong, and may be easily violated. 
For example, consider the simple three-variable case case with $\mu_1>\mu_2>\mu_3$. 
$X_1$ is independent of the other variables, but positive correlation exists between $X_2$ and $X_3$. 
Then whenever $X_3$ is observed to be relatively high, the conditional mean of $X_2$ is raised, perhaps higher than that of $X_1$. 

In practice, it is impossible to know whether observed values are unusually high or low relative to their mean. 
Therefore this assumption is impossible to verify.
Future work could explore conditions under which it holds. 
It is possible that this correlated test of conditional means is preferable to assuming independence, but further work is necessary to demonstrate its validity.
% unconditional mu1 > mu2 > mu3, with strong correlation btwn mu2 and mu3, \& less btwn mu1 and mu3. mu3 observed usually high → conditional mean of X2 raised, perhaps above that of X1.

\section{Additional SHAP Experiments}\label{apx:extra}

\subsection{StableSHAP Efficiency}

% Figure \ref{fig:boxplots} displays empirical error rates on top-$K$ Shapley rankings. 

We investigated how the adaptive allocation strategy of StableSHAP improved stability relative to Shapley Sampling.
To do so, we compared their performance given equal computational budgets. %explaining predictions of a neural network on the Adult dataset. 
The algorithms explained 30 test set predictions of a neural network trained on the Adult dataset.
On each input, we ran StableSHAP (Alg. \ref{alg:StableSHAP}) to convergence 50 times, tracking the average number of total samples $n=\sum_j n_j$ (Equation~\eqref{eq:Shapley Sampling}).
Then, we ran Shapley Sampling 50 times, evenly distributing the same total number of samples across the $d$ features for each input.
We repeated this procedure for $K=2$ and $K=5$, as well as $\alpha=0.1$ and $\alpha=0.2$.

\begin{figure}[ht]
    \centering
    \includegraphics[width=0.6\columnwidth]{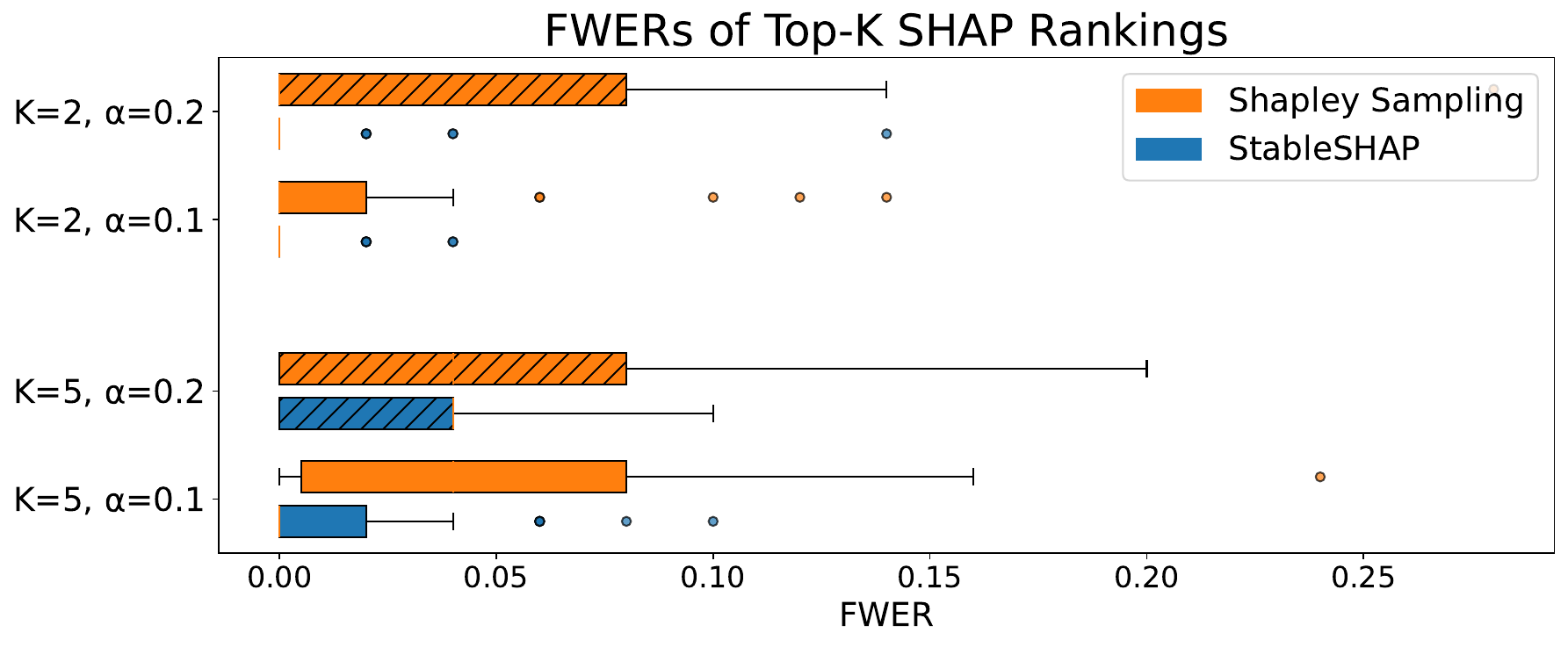}
    \caption{Stability of StableSHAP and Shapley Sampling, Equal Computation.}
    \label{fig:boxplots}
\end{figure}

Figure \ref{fig:boxplots} displays empirical error rates on top-$K$ Shapley rankings. 
For all input data points, ranks $K$, and tolerances $\alpha$, StableSHAP successfully achieved FWER control, as anticipated. 
In contrast, rankings from Shapley Sampling were considerably less stable. 
In all settings, their error rates were higher than StableSHAP's, often exceeding $\alpha$. 
This indicates that StableSHAP is more efficient than typical Shapley Sampling methods, which allocate samples equally across features.

\subsection{Variances}

\begin{figure}
    \centering
    \includegraphics[width=0.6\columnwidth]{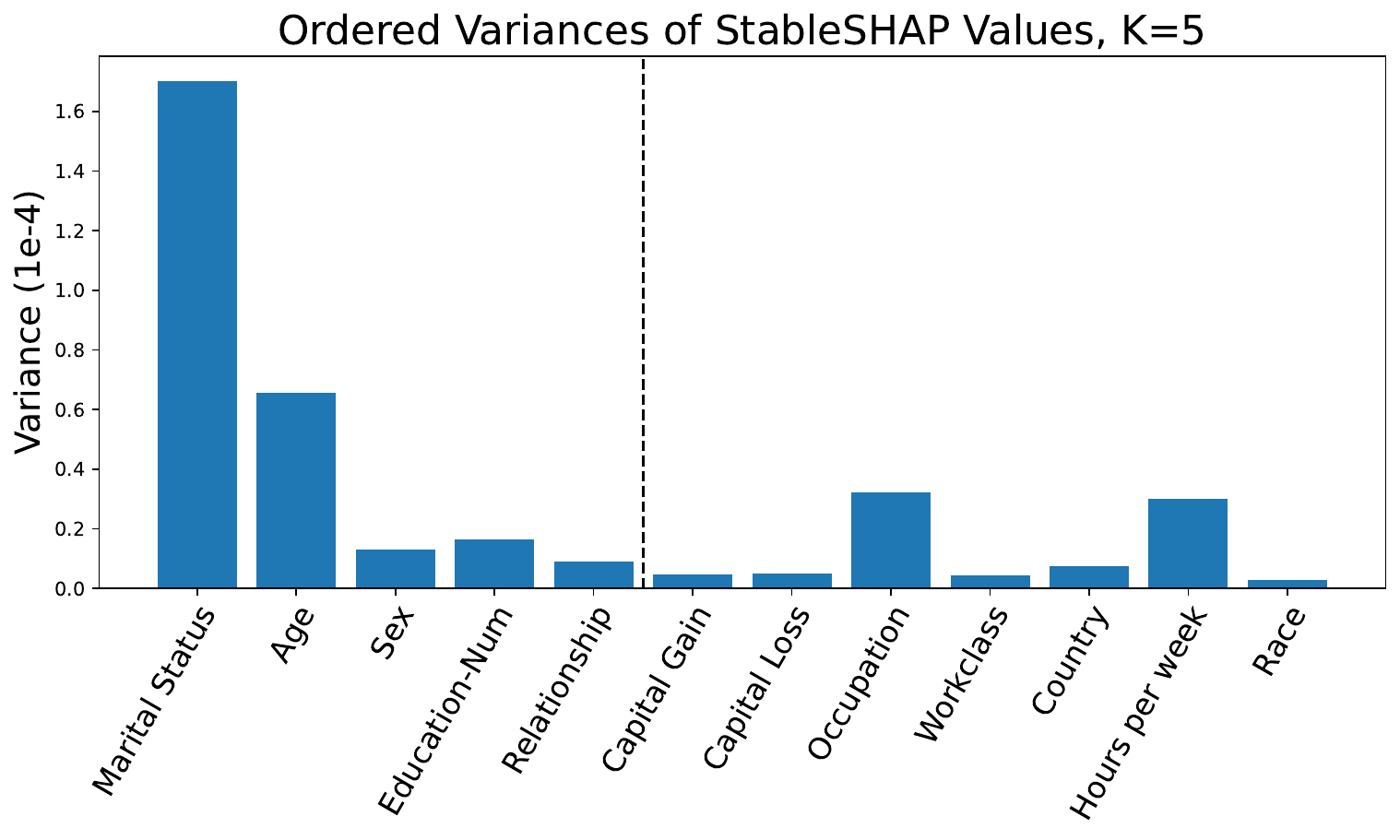}
    \caption{Variances of StableSHAP values displayed in Figure \ref{fig:perms_per_feature}}
    \label{fig:StableSHAP_vars}
\end{figure}

Figure \ref{fig:StableSHAP_vars} shows the variances of the SHAP estimates in Figure \ref{fig:perms_per_feature}. They follow the same order by magnitude, with dashed line at $K=5$. While lower SHAP values generally have smaller variance, the 8\textsuperscript{th} and 11\textsuperscript{th} ranks have high variance. This accounts for the iterations in Figure \ref{fig:perms_per_feature} for which Occupation and Hours of Week require more than 100 samples.

\section{Experimental Details}\label{apx:exp}

\begin{table*}[t]
\caption{Model Accuracies on Benchmark Datasets}
\label{tbl:AccuracyResults}
\centering
\begin{tabular}{l c}
\toprule
Dataset & Accuracy \\
\midrule
\texttt{Adult}        & 82.0\% \\
\texttt{Bank}         & 82.8\% \\
\texttt{BRCA}         & 77.0\% \\
\texttt{Credit}       & 72.0\% \\
\texttt{WBC}          & 91.8\% \\
\bottomrule
\end{tabular}
\end{table*}

For the models, we trained two-layer feedforward neural networks in Pytorch for 20 epochs. The hidden layer had 50 neurons. To discourage overfitting to the more common class, we batched the two classes in sampling each batch from the training data. We trained networks for 20 epochs in batches of 32 samples, using the Adam optimizer with a learning rate of 0.001. 

Datasets for experimental results were pulled from the UCI Machine Learning repository. We random split the data 75\%/25\% into training and test sets using \texttt{sklearn}. Accuracies of the five model are contained in Table \ref{tbl:AccuracyResults}

Computation was conducted using a slurm partition. Each entry in the tables was run with a separate job, though not with internal parallelization. In constructing Tables \ref{tbl:TopKResults} and \ref{tbl:LIMEResults}, we only considered input data points for which our algorithms successfully rejected all $K$ tests on at least 90\% of the iterations.

Our experiments treat the most common observed top-K as ground truth. In theory, this could be wrong. Fortunately, however, it is straightforward to show that the probability this surrogate ground truth is incorrect is essentially 0. 

To do so, recall that Table \ref{tbl:RetroResults} shows the most common top-K always occurs with frequency exceeding $1-\alpha$, according to the theory. To be conservative, suppose $\alpha=0.2$, and the most common top-$K$ appears $50\times (1-\alpha) = 40$ times. Now suppose the most common top-K is actually incorrect. Then the probability of observing the \textit{wrong} top-K at least 40 times is upper-bounded by the binomial probability
$$\mathbb{P}(\text{observe top-$K$ at least 40x}\ \vert\ \text{it is wrong}) \leq \sum_{s=40}^{50}\binom{n}{s} 0.2^s (1-0.2)^{50-s} \approx 1.29 \times 10^{-19}.$$
With this infinitesimal probability in mind, we can safely conclude that our ``proxy'' ground truth is indeed correct. Thus, synthetic experiments with exhaustively computed ground truth SHAP values is unnecessary. Doing so would require $\mathcal{O}(2^d)$ samples for each input, which is computationally prohibitive in high dimensions.

\subsection{SHAP}\label{apx:expSHAP}

Our experiments ranked the absolute values of SHAP estimates. 
The sum of all SHAP values for input $x$ is $f(x)-E[f(X)]$, which may be positive or negative. In the latter case, the most important features are the ones with the largest negative values. Further, an individual sample may have features with both large positive and negative SHAP values. An alternative approach would be flip the sign of the SHAP values if their sum is negative, then rank these preprocessed values. 

Because absolute values have non-negative support, their distribution is no longer Gaussian, but rather a folded normal. 
To avoid this, we used the same normal distribution, but flipped the sign if the SHAP estimate was negative. 
In theory, the true mean could then be the opposite sign as its observation.
In essence, this does not consider the distribution of the absolute SHAP values. 
Instead, it merely establishes a ranking based on their absolute value.
This subtle distinction may have virtually no practical difference, as the mass of the highest-ranking features may be almost entirely positive or negative.
% We are primarily interested in the highest-ranking features; their mass should not cross heavily between positive and negative.

Shapley Sampling  \eqref{eq:Shapley Sampling} takes the average of $n$ values of $v(S_j^i \cup \{j\}) - v(S_j^i)$. 
To select subsets $S_j^i \subseteq [d]\backslash\{j\}$, each iteration randomly permutes the $d$ features, then takes the features that precede $j$. 
This is formalized in Algorithm \ref{alg:SS}.

\begin{algorithm}%[tb]
\caption{Shapley Sampling}\label{alg:SS}
\begin{algorithmic}
\Require Input $x$, dataset $X$, Shapley value function $v(S)$, player $j \in [d]$, number of samples $n > 0$
\Ensure $\hat\phi_j(v)$, an unbiased estimate of $\phi_j(v)$
\State $\hat\phi_j \gets 0$
\For{$i=1$ {\bfseries to} $n$}
    \State $\pi_i \gets $ random permutation of $[d]$
    \State $S_j^i \gets$ elements of $\pi_i$ before $j$
    \State $\hat\phi_j \gets \hat\phi_j + v(S_j^i\cup\{j\}) - v(S_j^i)$
\EndFor
\State $\hat\phi_j(v) \gets \frac{1}{n}\hat\phi_j(v)$
\end{algorithmic}
\end{algorithm}

To generate Figure \ref{fig:instability}, we used results from the experiment generating Figure \ref{fig:boxplots}. 
This compared StableSHAP and Shapley Sampling on the Adult Census Income dataset. 
We chose an input for which Shapley Sampling was less stable. 
On the same input, we ran SPRT-SHAP, LIME, and the adjusted S-LIME procedure described in Appendix \ref{apx:lime}. 

Our SHAP estimates computed $v(S) = \hat\E[f(X)|X_S=x_s]$ with 10 samples per subset $S$. Features were sampled from their marginal distributions. 

We implemented KernelSHAP as described in \citet{CovertLee}. The same paper introduces a method for estimating its variance. However, \citet{controlSHAP} found that a considerably more stable approach was to compute a bootstrap estimate. This method takes the covariance matrix of Shapley estimates fit on bootstrapped versions of the data. In our experiments, we used this approach with 250 bootstrapped samples.

Our retrospective experiments (Table \ref{tbl:RetroResults}) used $2d+2048$ samples of subsets $S$. This is the default for KernelSHAP in the \texttt{shap} package. Note this is more than enough to guarantee convergence by CLT. Concerns of coarse approximation would only be legitimate under perhaps $n=30$ random samples. For each of 30 input data points, the empirical FWER was computed over 50 runs. The table displays the maximum FWER over these 30 inputs. 

Similarly, our top-$K$ experiments computed empirical FWERs over 50 runs, taking the maximum across 30 inputs.
NA values in Table \ref{tbl:TopKResults} correspond to settings of method, dataset, $K$, and $\alpha$ that proved incapable of converging with reasonable frequency.
Specifically, we ceased computation after our methods failed to converge on fewer than 10\% of inputs after at least 10 attempts.

StableSHAP used 100 initial samples per feature, and a maximum of 10,000. 
$n'_i$ and $n'_j$ were scaled according to their relative variances, following Equation~\eqref{eq:sample size unequal}. 
SPRT-SHAP iterations were capped at 50,000 samples.
The SPRT test was conducted every 1,000 iterations.
For the Type II error rate, we set $\beta=0.2$. 
This indicates the null will be accepted at most 20\% of the time. 
(In practice, the null is never accepted.)

StableSHAP and SPRT-SHAP do not necessarily use the same inputs in Table \ref{tbl:TopKResults}. 
To construct Figure \ref{fig:shap_n_samples}, we explicitly selected 30 inputs for which both top-$2$ ranking algorithms converged. 
For BRCA, there were not enough test set samples for which SPRT-SHAP stabilized the top-$2$ ranks under reasonable $\alpha$; therefore we used training samples on this dataset.

\subsection{LIME}\label{apx:lime}

Our LIME procedure can be implemented using Zhengze Zhou's \texttt{slime} repository almost entirely off-the-shelf. In our experiments, we added only 3 lines of code, which flag when the maximum number of samples have been used but not all hypothesis tests reject. 

S-LIME generates enough samples such that each highest-ranking feature beats its runner-up with probability exceeding $1-\alpha$, in a manner similar to Algorithm \ref{alg:StableSHAP}. 

To verify the winner, Test \eqref{eq:test} takes the highest p-value $p_{1j}$. 
Comparing to the runner-up is equivalent to a level-$\alpha/2$ Z-test.
Unless all variances are equal, ranks below the runner-up may have higher p-values \citep{fithian_hung}.
To run S-LIME in this context, it is necessary to assume the runner-up always has the highest p-value.

Making the requisite assumptions, running S-LIME at level $\alpha/2K$ for each test is therefore a valid level-$\alpha$ ranking procedure for LIME with $K$-Lasso.
The factors of 2 and $K$ account for selection and multiple testing, respectively.

To control FWER at level $\alpha$, the ``alpha'' parameter passed to \texttt{slime()} should be $\frac{\alpha}{2K}$. We used 1,000 initial samples, and a maximum of 100,000. Another parameter, ``tol,'' denotes the tolerance level of the hypothesis tests. Setting $tol=0$ corresponds to the algorithm in the paper, and is of course a viable option. 

The creators of the package found that having a small positive tolerance could yield comparable results with considerably greater efficiency. In our experiments, we set $tol=10^{-4}$, which ran more quickly while controlling the FWER at level $\alpha$. While higher values of \textit{tol} made S-LIME run more quickly, it also resulted in rare instances in which the algorithm allegedly converged without actually controlling the FWER. The arbitrary nature of this choice is a legitimate limitation of the S-LIME method. That said, our experimental errors are always below $\alpha$, and more a conservative approach could set $tol=0$.

\end{document}